\documentclass[journal ]{new-aiaa}
\usepackage[utf8]{inputenc}
\usepackage{textcomp}

\usepackage{enumitem}
\usepackage{graphicx}
\usepackage{amsmath}
\usepackage[version=4]{mhchem}
\usepackage{siunitx}
\usepackage{longtable,tabularx}
\usepackage{dsfont}
\usepackage{amsmath}
\usepackage{amsfonts}
\usepackage{algorithm}
\usepackage{algpseudocode}
\usepackage{booktabs}
\usepackage{multirow}
\usepackage{graphicx}

\newcommand{\vect}[1]{\boldsymbol{#1}}
\newcommand{\assump}[1]{\textbf{[#1]}}
\newcommand{\ez}{\varphi_{\mathrm{EZ}}}
\newcommand{\rr}{\varphi_{RR}}

\title{Inferring Turn-Rate-Limited Engagement Zones with Sacrificial Agents for Safe Trajectory Planning}

\author{Grant Stagg\footnote{PhD Candidate, Electrical and Computer Engineering, Brigham Young University.}}
\author{Cameron K. Peterson\footnote{Associate Professor, Electrical and Computer Engineering, Brigham Young University, and AIAA Senior Member.}}
\affil{Brigham Young University, Provo, Utah, 84602, USA}
\begin{document}

\maketitle

\begin{abstract}
This paper presents a learning-based framework for estimating pursuer parameters in turn-rate-limited pursuit--evasion scenarios using sacrificial agents. Each sacrificial agent follows a straight-line trajectory toward an adversary and reports whether it was intercepted or survived. These binary outcomes are related to the pursuer’s parameters through a geometric reachable-region (RR) model. Two formulations are introduced: a boundary-interception case, where capture occurs at the RR boundary, and an interior-interception case, which allows capture anywhere within it. The pursuer's parameters are inferred using a gradient-based multi-start optimization with custom loss functions tailored to each case. 

Two trajectory-selection strategies are proposed for the sacrificial agents: a geometric heuristic that maximizes the spread of expected interception points, and a Bayesian experimental-design method that maximizes the $D$-score of the expected Gauss--Newton information matrix, thereby selecting trajectories that yield maximal information gain. Monte Carlo experiments demonstrate accurate parameter recovery with five to twelve sacrificial agents. The learned engagement models are then used to generate safe, time-optimal paths for high-value agents that avoid all feasible pursuer engagement regions.
\end{abstract}

\section{Introduction}
\lettrine{I}{n} modern warfare environments, autonomous agents are tasked with safely navigating airspace while avoiding adversarial threats that actively seek to neutralize them. These environments are often characterized by partial or incomplete information about threat locations and capabilities, including speed, turn rate, and engagement range. Effective operation therefore requires methods that can infer adversary parameters from limited observations while explicitly accounting for uncertainty.

Crucially, this inferred information must be incorporated immediately into decision-making. Autonomous agents must plan trajectories that avoid regions where interception is likely, even as uncertainty in adversary capabilities persists.

A natural framework for reasoning about such interactions is the concept of an \emph{engagement zone} (EZ): the set of states or spatial regions in which an adversary can successfully intercept a target, given its kinematic and operational constraints. EZs provide a principled representation of threat capabilities and enable quantitative assessment of the risk associated with different portions of the environment.

However, the parameters defining these EZs (pursuer position, heading, turn-rate, range, and speed) are often unknown. To address this, we consider the use of sacrificial agents, which deliberately expose themselves to potential engagement. By observing the outcomes of these interactions, whether the agent is intercepted or survives, we can infer the underlying pursuer parameters and, in turn, construct a probabilistic model of the EZ.

In this paper, we present a framework for inferring pursuer parameters from the outcomes of sacrificial agent interactions. The approach defines loss functions that map interception and survival outcomes to a likelihood over pursuer parameters, and employs gradient-based optimization to recover candidate estimates of the pursuer’s position, heading, turn radius, range, and speed. To mitigate sensitivity to local minima, a multi-start optimization procedure is used, yielding a set of feasible pursuer parameter vectors consistent with the observed outcomes. 

Building on this inference framework, we design a sacrificial agent path planner that deliberately selects informative, high-risk trajectories to maximize information gain about the adversary. Finally, we demonstrate how this learned set of potential pursuer parameter vectors can be directly incorporated into motion planning, enabling autonomous agents to generate trajectories that avoid all engagement zones induced by the inferred uncertainty.

The contributions of this work are as follows:
\begin{enumerate}
\item An optimization-based, multi-start inference framework for estimating pursuer parameter vectors from sacrificial-agent trajectories and binary interception outcomes, producing a set of pursuer parameter vectors consistent with observed data.
\item Two interception models and corresponding differentiable loss functions that enable gradient-based estimation of pursuer parameters under different assumptions about when interception occurs.
\item Two sacrificial-trajectory selection strategies, one based on geometric exploration and one based on information-driven design, that improve the efficiency of estimating the set of pursuer parameter vectors.
\item A trajectory-planning formulation that uses the inferred set of pursuer parameter vectors to generate safe, time-optimal paths that avoid all engagement zones consistent with the inferred parameters.
\end{enumerate}

More broadly, this work studies how autonomous agents can acquire information about adversarial capabilities through interaction when observations are limited to binary outcomes. Rather than relying on continuous sensing or direct measurement, only whether an agent is intercepted or survives is measured. We show that these sparse observations are sufficient to constrain adversary parameters and that sacrificial agent paths can be planned to efficiently reduce uncertainty and support subsequent safe trajectory planning.

The remainder of this paper is organized as follows. Section~\ref{sec:related_work} reviews related work, and Section~\ref{sec:background} provides background on EZs. Section~\ref{sec:problem} formulates the problem, while Section~\ref{sec:param_learning} presents the pursuer-parameter learning algorithm. The sacrificial-trajectory planning approach is described in Section~\ref{sec:sacrificial_path}.  Once the feasible pursuer parameters are learned, Section~\ref{sec:safe_path} demonstrates how to incorporate them into a safe trajectory planner. Section~\ref{sec:results} presents simulation results, and conclusions are given in Section~\ref{sec:conclusion}.

\subsection{Related Work}\label{sec:related_work}
Engagement zones (EZs) provide a principled framework for representing the set of states from which an adversary can intercept a target. Early work modeled EZs using simplified cardioid geometries~\cite{weintraub2022optimal,dillon2023optimal,wolek2024sampling}, which capture interception risk but are not directly derived from the pursuer’s physical parameters. More recent work introduced basic engagement zones (BEZs) derived from the geometric structure of differential games~\cite{von2023basic,isaacs1965differential}, yielding engagement regions that explicitly encode pursuer–evader dynamics and agent capabilities.

The BEZ formulation assumes the pursuer can instantaneously reorient to an optimal collision course. This assumption was relaxed by introducing a turn-rate-limited engagement zone, the curve–straight basic engagement zone (CSBEZ), which models pursuer motion using Dubins kinematics~\cite{dubins1957curves,chapman2025engagement}. While BEZ and CSBEZ models provide a stronger geometric and theoretical foundation than cardioid approximations, they generally assume perfect knowledge of the pursuer’s parameters, including position, heading, speed, turn radius, and range.

In realistic adversarial settings, such information is rarely known with certainty. This limitation was partially addressed in~\cite{chapman2025safe}, where varying levels of knowledge were assumed to plan safe paths. In our prior work, we extended BEZs to probabilistic settings~\cite{stagg2025acc} and later to turn-rate-limited pursuers~\cite{Stagg25_CSPEZ}, producing uncertainty-aware EZ boundaries. These methods, however, still required an initial probability distribution over the pursuer’s parameters. The work presented here removes that dependency by using sacrificial (decoy) agents deployed to infer the pursuer’s parameters directly.

Prior work has explored the use of sacrificial agents—low-cost platforms that deliberately undertake risky actions to acquire valuable information. In our prior work, this paradigm was applied to radar mapping, where expendable agents intentionally exposed themselves to detection in order to localize and characterize enemy radar systems, trading survival for improved situational awareness~\cite{Stagg25_RADAR}. Related ideas appear in the path-clearance problem~\cite{path_clearance,pathclearance_info}, where scouting agents are deployed ahead of high-value assets to identify hidden threats. In~\cite{cesare2015multi}, agents nearing battery depletion are repurposed as sacrificial scouts, while Duggan et al.~\cite{duggan2025uncertaintyawareplanningheterogeneousrobot} employ fast, low-cost agents to probe uncertain terrain and secure safe routes for more valuable teammates.

In contrast to prior work, we use sacrificial agents that are expected to be intercepted in order to infer adversary capabilities. While sacrificial agents provide a mechanism for collecting informative interaction data, learning the underlying engagement geometry from such data requires new approaches for constructing reachable regions (RRs) and engagement zones (EZs) directly from observations.

Prior work on learning reachable regions (RRs) and engagement zones (EZs) has primarily relied on simulation-based approaches, where interception probabilities are estimated through repeated Monte Carlo simulations of pursuer–evader interactions~\cite{dantas2021weapon,dantas2023realtimesurfacetoairmissileengagement}. Candidate evader states or trajectories are evaluated under assumed pursuer models, and empirical interception probabilities are computed as the fraction of simulations resulting in capture. EZs are then defined implicitly as level sets of the resulting probability field.

While these methods are data-driven, the engagement geometry itself is not inferred directly from observed interactions. Instead, it is approximated via forward simulation of hypothesized pursuer models. In contrast, our approach leverages real sacrificial agent trajectories as direct evidence of the underlying engagement structure, enabling EZs to be inferred from observed interception and survival outcomes rather than simulated engagements.

% section about baysian experimental design
Our sacrificial trajectory selection is closely related to Bayesian experimental design (BED), in which actions are chosen to maximize expected information about uncertain model parameters. Many BED approaches use the Fisher Information Matrix (FIM) as a design criterion, commonly maximizing D-optimality (log-determinant of the FIM) or E-optimality (minimum eigenvalue). Trajectory optimization methods have explicitly maximized Fisher information~\cite{wilson2015maximizing}, and observer path planning problems have used D-optimality to guide trajectory selection~\cite{kaya2022observer}. In Bayesian settings, maximizing the expected trace or determinant of the FIM provides tractable surrogates for information gain~\cite{overstall2022properties,prangle2023bayesian}, closely related to Gauss–Newton information approximations used in nonlinear regression. Related work has also employed mutual information and Bayesian optimization for informative path planning~\cite{marchant2014bayesian,francis2019occupancy,morere2017sequential}.

While BED has been widely applied to active sensing and experimental design, it has not previously been integrated into engagement-zone learning for adversarial interactions. In contrast, the framework proposed here couples pursuer-parameter inference with BED-based sacrificial trajectory selection, enabling agents to actively probe and learn adversary capabilities. The resulting set of pursuer parameter vectors captures model uncertainty and directly supports downstream planning, as demonstrated in Section~\ref{sec:safe_path}, where it is used to generate safe, time-optimal trajectories that avoid all learned engagement zones.

\subsection{Background: Weapon Engagement Zones}\label{sec:background}
\begin{figure}[t]
    \centering
    \includegraphics[]{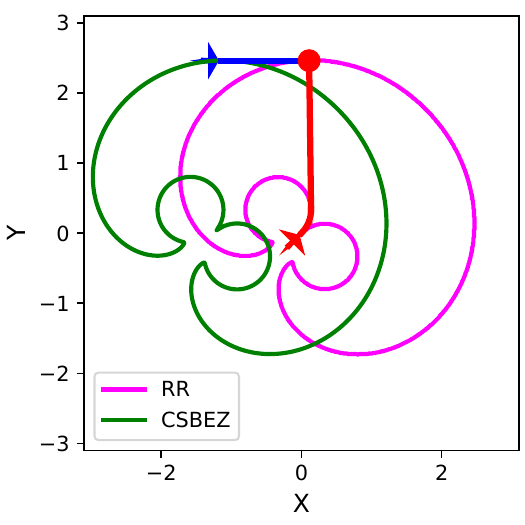}
\caption{Reachable region (RR) of the pursuer (magenta) and the corresponding CSBEZ (green) for an evader heading along the $x$-axis. Sample evader (blue) and pursuer (red) trajectories are also shown.}
    \label{fig:CSBEZ}
\end{figure}

The curve–straight basic engagement zone (CSBEZ) characterizes the set of evader configurations—positions and headings—from which interception is possible by a pursuer with bounded turn rate (minimum turn radius $a$). The model assumes the evader maintains its initial heading at constant speed (see the green region in Fig.~\ref{fig:CSBEZ}), while the pursuer follows an optimal turn–straight collision course.

This region depends on the pursuer’s initial state and parameters, collected into the pursuer parameter vector
\begin{equation}
\vect{\theta}_P = (\vect{x}_P, \psi_P, v_P, a, R),
\end{equation}
where $\vect{x}_P \in \mathbb{R}^2$ is the pursuer’s initial position, $\psi_P$ its heading, $v_P$ its speed, $a$ its minimum turn radius, and $R$ its engagement range. The CSBEZ also depends on the evader’s initial position $\vect{x}_E$, heading $\psi_E$, and speed $v_E$.

To evaluate whether a given evader configuration lies inside the CSBEZ, we project the evader's future position assuming it travels in a straight line for the time it takes the pursuer to traverse its maximum range $R$ at constant speed $v_P$. This projected point is defined as
\begin{equation}
    \vect{x}_F = \vect{x}_E + \nu R \vect{v}_E,
\end{equation}
where $\nu = v_E / v_P$ is the speed ratio and $\vect{v}_E = [\cos \psi_E, \sin \psi_E]^\top$ is the unit vector in the direction of the evader’s heading. The pursuer attempts to reach this point via the shortest feasible curve-straight path, which consists of a circular arc of radius $a$, either a left or right turn, followed by a straight-line segment. The engagement zone function computes the length of this path and subtracts the pursuer's range:
\begin{equation}
    \ez
    (\vect{x}_E, \psi_E, \vect{\theta}_P) = L(\vect{x}_F, \vect{x}_P, \psi_P, a) - R.
    \label{eq:ez_function}
\end{equation}
A point lies inside the CSBEZ if $\ez(\vect{x}_E, \psi_E, \vect{\theta}_P) \leq 0$, where negative values indicate that the evader lies within the interceptable region.

The function \( L(\vect{x}_F, \vect{x}_P, \psi_P, a) \) returns the shortest total length of either a left-turn–straight or right-turn–straight path from $\vect{x}_P$ to $\vect{x}_F$, assuming an initial heading of $\psi_P$ and a minimum turn radius $a$. That is,
\begin{equation}
    L(\vect{x}_F, \vect{x}_P, \psi_P, a) = 
    \min\big(L_\ell(\vect{x}_F, \vect{x}_P, \psi_P, a),\, L_r(\vect{x}_F, \vect{x}_P, \psi_P, a)\big),
\end{equation}
where \( L_\ell \) and \( L_r \) are the lengths of the left- and right-turn paths, respectively. These constructions follow from the geometric derivation presented in~\cite{Stagg25_CSPEZ}.

We define the reachable region (RR) function similarly, but without reference to any evader state. It evaluates the distance required for the pursuer to reach a fixed point $\vect{x} \in \mathbb{R}^2$ using a curve-straight path from its initial state. This function is independent of the evader's position, heading, or speed, and is given by
\begin{equation}
    \rr(\vect{x}, \vect{\theta}_P) = L(\vect{x}, \vect{x}_P, \psi_P, a) - R.
    \label{eq:rs_function}
\end{equation}
Thus, $\rr(\vect{x}, \vect{\theta}_P)$ computes the curve-straight path length from $\vect{x}_P$ to $\vect{x}$ and subtracts the pursuer's available range $R$. The boundary of the reachable region corresponds to the zero contour of $\rr(\vect{x}, \vect{\theta}_P) = 0$.

The EZ function can be viewed as a shifted version of the RR function, where the target point $\vect{x}$ is displaced along the evader’s heading to account for its future position at time $R / v_P$. This shift introduces a dependence on the evader’s state, distinguishing the EZ from the purely kinematic RR. Figure~\ref{fig:CSBEZ} shows an example of an RR (magenta) and EZ (green) for a set of pursuer and evader parameters.

\section{Problem Statement}\label{sec:problem}

To better understand what can be learned about unknown adversaries using sacrificial vehicles, we examine the inference problem under a range of assumptions and parameters.  
We present a hierarchy of learning cases with increasing levels of complexity to illustrate what can be inferred about the pursuer. Case~1 considers the setting in which the engagement zone shape is known and the goal is to estimate only the pursuer's position and heading. Case~2 increases the difficulty by also treating the shape (defined from the turn radius and range) as unknown. Case~3 is the most general setting, in which pursuer speed is also unknown, and an additional timing measurement is introduced to make speed observable. For each of these three case families, we consider both a noise-free variant (A) and a noisy variant (B), where trajectory measurements are corrupted by additive noise.

Given a collection of $N_s$ sacrificial agent trajectories 
$\mathcal{X} = \{\, \vect{x}_s^{(i)}(t) \,\}_{i=1}^{N_s}$ 
and corresponding binary outcomes 
$\mathcal{Z} = \{\, z^{(i)} \,\}_{i=1}^{N_s}$, 
where $z^{(i)} \in \{0, 1\}$ indicates whether the $i^\text{th}$ agent was intercepted ($z^{(i)}=1$) or survived ($z^{(i)}=0$), 
we aim to infer the parameters of an unknown pursuer. 
Each trajectory $\vect{x}_s^{(i)}(t)$, defined for $t \in [t_0^{(i)}, t_f^{(i)}]$, 
terminates at time $t_f^{(i)}$ and position 
$\vect{x}_s^{(i)}(t_f^{(i)})$. 
This final position is either where the agent was intercepted or when it exits the region of interest $\mathcal{D} \subset \mathbb{R}^2$.

Each trajectory corresponds to a straight inbound path toward the pursuer, parameterized by its starting location $\vect{x}_s^{(i)}(t_0^{(i)})$ and heading $\psi_s^{(i)}$. 
For clarity, we treat each agent as beginning directly from $\vect{x}_s^{(i)}(t_0^{(i)})$; 
in practice, an agent would depart from a common origin, travel to this start point, and then execute the inbound segment. 
We discard the transit leg because it remains outside the pursuer’s reachable region and therefore carries no information about capture feasibility.
The resulting collection $(\mathcal{X}, \mathcal{Z})$ constitutes the dataset used to infer the pursuer’s parameter vector~$\vect{\theta}_P$.

The unknown pursuer parameters are denoted:
\begin{equation}\label{eq:thetaP}
 \vect{\theta}_P = [\, x_P,\; y_P,\; \psi_P,\; a,\; R,\; v_P \,]^\top \in \mathbb{R}^6,
\end{equation}
where $(x_P, y_P)$ is the initial (launch) position of the pursuer, $\psi_P$ is its heading, $a$ is its minimum turn radius, $R$ is its maximum range, and $v_P$ is its constant speed. 

\subsection{Assumptions}

The operational assumptions used in our analysis are listed below. They form the modeling framework for inferring engagement zones.
\begin{itemize}
    \item \assump{A1-strict} \textit{Boundary interception:} Interception occurs exactly on the boundary of the pursuer's reachable region.
    \item \assump{A1-relaxed} \textit{Interior interception:} Interception may occur anywhere within the pursuer's reachable region.
    \item \assump{A2} \textit{Guaranteed interception:} If the pursuer is able to intercept the evader, it does so.
    \item \assump{A3} \textit{Stationary launch point:} The pursuer's launch position \((x_P, y_P)\), and heading ($\psi_P)$ are fixed.
    \item \assump{A4} \textit{Known evader behavior:} Each sacrificial agent follows a known, constant-velocity trajectory \(\vect{x}_s^{(i)}(t)\).
    \item \assump{A5} \textit{Kinematic-only model:} Engagement zones are defined purely by kinematic constraints; effects such as terrain, occlusion, or probabilistic detection are ignored.
    \item \assump{A6} \textit{Multi-target capability:} The pursuer is capable of engaging multiple evaders; for example, it may represent a missile launch system with multiple interceptors.
\end{itemize}
These assumptions define the analytical regime under which inference is performed. Assumptions \assump{A1-strict} and \assump{A1-relaxed} specify alternative interception models, enabling evaluation under both boundary-based and interior-inclusive interpretations of engagement. Assumptions \assump{A3} and \assump{A4} fix the pursuer configuration and prescribe sacrificial agent trajectories, which is consistent with experimental settings where agent paths are controlled or planned. Assumption \assump{A5} restricts attention to kinematic effects, allowing the inferred engagement zones to reflect pursuer capability rather than environmental or sensing artifacts. Finally, assumption \assump{A6} permits multiple sacrificial trials to be interpreted as independent observations of a single pursuer configuration, rather than requiring reinitialization of the system after each engagement; a similar assumption is adopted in~\cite{chapman2025safe}.

% These assumptions also indicate directions for future extensions of our approach. For instance, \assump{A3}, which assumes a stationary launch point, is realistic for certain adversarial systems but does not account for mobile systems, which represent an important and interesting direction for further development. As this work presents the first framework capable of inferring engagement zones, we focus here on establishing that framework and leave the incorporation of additional conditions that relax these assumptions to future research.

Several of these assumptions also identify natural extensions of the framework. For example, \assump{A3} is appropriate for fixed-site systems but does not capture mobile pursuers, which represents a clear direction for future work. More broadly, the focus of this paper is to establish a principled framework for engagement zone inference under well-defined conditions; extensions that relax these assumptions can be incorporated within the same structure in subsequent works.

\subsection{Measurements}
The inference algorithm relies on a small number of measurable quantities collected from each sacrificial trajectory: position history, interception outcome, and, in some cases, pursuer launch timing. 
For each sacrificial agent, the following data are available:
\begin{itemize}
    \item \textbf{Trajectory:} The full trajectory \(\vect{x}_s^{(i)}(t)\) for \(t \in [t_0^{(i)}, t_f^{(i)}]\). In uncertainty cases, the observed trajectory is modeled as the true path corrupted by additive Gaussian noise with zero mean and covariance~$\Sigma_{\vect{x}}$.
    \item \textbf{Interception outcome:} A binary indicator \(z^{(i)} \in \{0, 1\}\) specifying whether the agent was intercepted (1) or survived (0).
    \item \textbf{Time of launch (speed-unknown setting only):} The pursuer's launch time \(t_{\ell}^{(i)}\), which is used when pursuer speed \(v_P\) is treated as unknown. This measurement is also corrupted by additive Gaussian noise with zero mean and variance $\sigma_t^2$.
\end{itemize}
The agent’s trajectory \(\vect{x}_s^{(i)}(t)\) is assumed to be measured using onboard sensors such as inertial measurement units and, when available, GPS or other localization systems. The details of the navigation system are outside the scope of this work; instead, navigation uncertainty is explicitly modeled in the noisy variants (the B use cases) by introducing additive noise.
In the speed-unknown setting, the launch time \(t_{\ell}^{(i)}\) of the pursuer is assumed to be known. In practice, this value could be estimated using external sensing modalities such as acoustic or visual sensors. For example, an acoustic sensor might detect the sudden sound of a missile launch, while a visual system (e.g., high-speed camera or infrared sensor) could identify the moment of ignition or departure.

In many settings, only trajectories and binary outcomes are available; in this measurement regime, the pursuer speed is not identifiable from capture feasibility alone because timing can be rescaled without changing interception outcomes. Accordingly, learning $v_P$ requires an additional timing cue (e.g., launch time), which is only assumed available in the speed-unknown setting.

\subsection{Learning Cases}
We consider a hierarchy of learning cases with increasing complexity. Each case specifies (i) which elements of \(\vect{\theta}_P\) are unknown, (ii) whether sacrificial agent trajectory measurements are corrupted by noise, and (iii) what measurements are available. All cases are evaluated under both the boundary-only interception assumption (\assump{A1-strict}) and the interior-inclusive interception assumption (\assump{A1-relaxed}).

Cases~1 and~2 assume that the pursuer's speed is known. Case~1 further assumes that the pursuer's kinematic parameters are known, and isolates the localization problem by estimating only the pursuer's position and heading. Case~2 relaxes this assumption by treating the kinematic limits (e.g., turn radius and range) as unknown while still assuming known speed.

Together, these cases function as an ablation across sensing regimes: Cases~1--2 examine what can be inferred from interception outcomes and trajectory information alone, without access to timing measurements, whereas Case~3 augments the measurement set with launch-time information, rendering the pursuer speed observable.

Table~\ref{tab:learning_cases} summarizes the learning cases considered in this work.

\begin{table*}[ht]
\centering
\caption{Learning cases for pursuer parameter estimation. Each case is evaluated under both the boundary-only interception assumption (\assump{A1-strict}) and the interior-inclusive interception assumption (\assump{A1-relaxed}).}
\label{tab:learning_cases}
\begin{tabular}{@{} >{\centering\arraybackslash}p{1.0cm} p{3.7cm} >{\centering\arraybackslash}p{1.9cm} p{4.3cm} p{5.4cm}@{}}
\toprule
\textbf{Case} & \textbf{Unknown Parameters} & \textbf{Uncertainty} & \textbf{Measured Data} & \textbf{Description} \\
\midrule
1A & Position \((x_P, y_P)\), heading \(\psi_P\) & No  & Agent trajectory, interception outcome & Known engagement zone shape; estimate pursuer position and heading. \\
1B & Position \((x_P, y_P)\), heading \(\psi_P\) & Yes & Noisy agent trajectory, interception outcome & Same as 1A, but with Gaussian noise corrupting trajectory measurements. \\
2A & \((x_P, y_P), \psi_P\), turn radius \(a\), range \(R\) & No  & Agent trajectory, interception outcome & Engagement zone shape unknown; estimate position, heading, turn radius, and range. \\
2B & \((x_P, y_P), \psi_P\), \(a\), \(R\) & Yes & Noisy agent trajectory, interception outcome & Same as 2A, but with trajectory corrupted by noise. \\
3A & Full \(\vect{\theta}_P\): \((x_P, y_P, \psi_P, v_P, a, R)\) & No  & Agent trajectory, interception outcome, time of launch & General case; estimate all pursuer parameters including speed. \\
3B & Full \(\vect{\vect{\theta}}_P\) & Yes & Noisy agent trajectory, interception outcome, time of launch & Most general case; full parameter inference with measurement uncertainty, including unknown pursuer speed. \\
\bottomrule
\end{tabular}
\end{table*}

To formalize which parameters are inferred in each case, we decompose the pursuer parameter vector as
\begin{equation}
 \vect{\theta}_P =
\begin{bmatrix}
 \vect{\theta}_P^S \\
 \vect{\theta}_P^{*\setminus S}
\end{bmatrix},
\end{equation}
where \(\vect{\theta}_P^S\) denotes the subset of parameters being learned and \(\vect{\theta}_P^{*\setminus S}\) denotes parameters assumed known.

In Case~1, only the pursuer's position and heading are unknown,
\begin{equation}
 \vect{\theta}_P^S = [x_P, y_P, \psi_P]^\top,
\end{equation}
while the turn radius \(a\), range \(R\), and speed \(v_P\) are assumed known. This case isolates the geometric localization problem under a known engagement zone shape.

In Case~2, the engagement zone shape itself is unknown. The learned parameter set expands to
\begin{equation}
 \vect{\theta}_P^S = [x_P, y_P, \psi_P, a, R]^\top,
\end{equation}
while the pursuer speed \(v_P\) remains known. This case captures the setting where both the pursuer state and kinematic limits must be inferred.

In Case~3, the full pursuer parameter vector is estimated:
\begin{equation}
 \vect{\theta}_P^S = [x_P, y_P, \psi_P, a, R, v_P]^\top.
\end{equation}
This represents the most general setting considered, in which all pursuer parameters are unknown and must be inferred jointly from interception outcomes and trajectory observations.

For each of the above parameter configurations, we additionally consider both noise-free (A) and noisy (B) measurement variants, where the latter assumes additive Gaussian noise corrupting the observed sacrificial agent trajectories. This structure enables a systematic evaluation of inference performance as uncertainty and problem complexity increase.

\section{Learning CSBEZ Parameters}\label{sec:param_learning}

The goal of the learning algorithm is to identify the pursuer parameter vector that best explain the sacrificial agent trajectories \(\vect{x}^{(i)}(t)\), the binary outcomes \(z^{(i)}\), and, for cases 3A and 3B, the launch times \(t_{\ell}^{(i)}\). 
We use a gradient-based optimization algorithm with a loss function designed to penalize infeasible pursuer parameter vectors. When the number of sacrificial trajectories is limited, multiple feasible parameter vectors may explain the data. To address this ambiguity, we employ a multi-start approach, running the optimization from many different initial guesses. This produces a set of candidate pursuer parameter vectors that all satisfy the observed outcomes.
To evaluate convergence, we estimate the variability of the resulting set of parameter vectors by computing the standard deviation across the optimized solutions.

The loss functions are designed to satisfy the following properties: 1) they are zero when evaluated at the true pursuer parameter vector, 2) they are strictly positive for infeasible pursuer parameter vectors.

This section proceeds as follows: the loss function for the boundary interception \assump{A1-strict} is described in Section~\ref{sec:loss_function_strict}, the loss function for the interior interception \assump{A1-relaxed} is shown in Section~\ref{sec:loss_function_relaxed}, the optimization algorithm is detailed in Section~\ref{sec:optimization_strict}, and the multi-start scheme is presented in Section~\ref{sec:particle_strict}.

\subsection{Loss Function: Boundary Interception}\label{sec:loss_function_strict}
We use the same loss function for cases 1A and 2A, with slight modifications in cases 1B and 2B to account for noise in the observations. 
In Cases 3A and 3B, estimating pursuer speed requires adding a term that incorporates the measured launch time. Without timing information, pursuer speed is not identifiable because speed changes can be offset by corresponding changes in time without altering interception feasibility. 
Therefore, in Case~3 we introduce an additional loss term to account for this timing measurement and to enable joint estimation of speed alongside the remaining pursuer parameters.

In Cases~1 and~2, the pursuer's speed is assumed to be known. In practice, this assumption is often reasonable: speed may be available from prior intelligence regarding the adversary system, or estimated through a brief tracking phase before sacrificial trajectories are deployed.

\subsubsection{Known Pursuer Speed (Cases 1 and 2)}
In this subsection, we describe the loss function for cases when the pursuer's speed is known.
We use a squared hinge loss defined as
\begin{equation}
    \mathrm{ReSq}(x)=\frac{1}{2}\max(0,x)^2,
\end{equation}
which provides a differentiable objective that penalizes parameter values inconsistent with the observed interception events.

The loss terms are defined as functions of the sacrificial trajectory $\vect{x}_s^{(i)}(t)$, the interception point $\vect{x}_s^{(i)}(t_f)$, the observed outcome $z^{(i)}$, and the pursuer parameter vector $\vect{\theta}_P$. The objective function is designed to enforce three conditions derived from the problem assumptions:
\begin{enumerate}
    \item The interception point lies on the boundary of the RR.
    \item Prior to interception, the trajectory remains outside the RR.
    \item A trajectory corresponding to a non-intercepted agent remains outside the RR at all times.
\end{enumerate}
Together, these conditions encourage consistency between the predicted RR and the observed interception outcomes across all sacrificial trajectories.

The first condition is enforced through a boundary-consistency loss,
\begin{equation}\label{eq:RR_bound_loss}
\mathcal{L}_{\text{bound,RR}}^{(i)}(\vect{\theta}_P,\vect{x}_s^{(i)}(t),z^{(i)})=
\begin{cases}
0, & z^{(i)}=0,\\[3pt]
\mathrm{ReSq}\!\big(\rr(\vect{x}_s^{(i)}(t_f),\vect{\theta}_P)-\varepsilon\big)
+\mathrm{ReSq}\!\big(-\rr(\vect{x}_s^{(i)}(t_f),\vect{\theta}_P)-\varepsilon\big), & z^{(i)}=1,
\end{cases}
\end{equation}
where $\varepsilon$ is a soft-margin factor that accounts for measurement noise. We define
\[
\varepsilon = \beta\,\sqrt{\lambda_{\max}(\Sigma_{\vect{x}})},
\]
where $\lambda_{\max}(\Sigma_{\vect{x}})$ is the largest eigenvalue of the position covariance $\Sigma_{\vect{x}}$, and $\beta\!\ge\!0$ controls the margin size. With this definition, any model whose predicted interception point lies within a distance $\varepsilon$ of the RR boundary incurs zero loss; the loss becomes positive only when the point lies more than $\varepsilon$ from the boundary. The function $\rr(\vect{x},\vect{\theta}_P)$ acts as a signed distance to the predicted RR boundary (positive outside, negative inside), so the two $\mathrm{ReSq}(\cdot)$ terms penalize deviations on either side of the boundary beyond the $\varepsilon$-margin. This loss is applied only when $z^{(i)}=1$ (intercepted agents); for non-intercepted trajectories, boundary consistency is not enforced since no terminal contact point is observed.

The second and third conditions are enforced through a trajectory-level loss that penalizes any portion of a sacrificial trajectory that lies within the predicted RR:
\begin{equation}
\mathcal{L}_{\text{traj,RR}}^{(i)}(\vect{\theta}_P,\vect{x}_s^{(i)}(t))
=\max_{t\in\vect{t}}
\Big[\mathrm{ReSq}\!\big(-\rr(\vect{x}_s^{(i)}(t),\vect{\theta}_P)-\varepsilon\big)\Big],
\end{equation}
where $\vect{t}=\{t_0,\,t_0+\Delta,\,\ldots,\,t_f-\alpha\}$ denotes the set of sampled time points along the trajectory, $\Delta=(t_f-\alpha-t_0)/N_t$, $N_t$ is the number of samples, and $\alpha>0$ offsets the final time to exclude points immediately preceding interception.

The argument of the $\mathrm{ReSq}(\cdot)$ becomes positive when a trajectory point lies more than $\varepsilon$ inside the predicted RR, indicating a violation of consistency between the parameter vector and the observed trajectory. When all trajectory points remain outside the RR by at least $\varepsilon$, this term evaluates to zero. The outer $\max(\cdot)$ selects the largest such violation along the trajectory, ensuring that even a single incursion into the RR produces a nonzero penalty.

Operationally, this loss discourages candidate parameter vectors $\vect{\theta}_P$ that produce an RR overlapping observed sacrificial trajectories, for both intercepted and non-intercepted agents. This prevents solutions in which the RR expands to explain interceptions at the expense of incorrectly overlapping with trajectories that were observed to survive.

The total loss for a single sacrificial agent is the sum of the boundary and trajectory components,
\begin{equation}
\mathcal{L}_{\text{comb,RR}}^{(i)}(\vect{\theta}_P,\vect{x}_s^{(i)}(t),z^{(i)})
=\mathcal{L}_{\text{bound,RR}}^{(i)}(\vect{\theta}_P,\vect{x}_s^{(i)}(t),z^{(i)})
+\mathcal{L}_{\text{traj,RR}}^{(i)}(\vect{\theta}_P,\vect{x}_s^{(i)}(t)).
\end{equation}
Finally, the losses from all sacrificial agents are summed to form the overall objective,
\begin{equation}\label{eq:rs_total_loss}
\mathcal{L}_{\text{tot,RR}}(\vect{\theta}_P,\mathcal{X},\mathcal{Z})
=\sum_{\vect{x}_s^{(i)}\in\mathcal{X},\,z^{(i)}\in\mathcal{Z}}
\mathcal{L}_{\text{comb,RR}}^{(i)}(\vect{\theta}_P,\vect{x}_s^{(i)}(t),z^{(i)}),
\end{equation}
where $\mathcal{X}=\{\vect{x}_s^{1}(t),\ldots,\vect{x}_s^{N_s}(t)\}$ is the set of all trajectories and $\mathcal{Z}=\{z^{1},\ldots,z^{N_s}\}$ the corresponding outcomes. This formulation jointly enforces geometric consistency between predicted RR boundaries and observed interception outcomes, while penalizing trajectories that violate the spatial constraints implied by the RR model. The result is a differentiable loss that aligns learned pursuer parameters with physically consistent engagement behavior.

\subsubsection{Unknown Pursuer Speed (Case 3)}
Using only trajectory and outcome measurements, the pursuer’s speed is unobservable. 
Inferring speed requires at least one additional measurement. 
In particular, incorporating launch-time information (e.g., from acoustic or visual sensors) provides the temporal reference needed for the algorithm to learn the speed. 
However, the loss function in Equation~\eqref{eq:rs_total_loss} does not enable speed to be learned. 
Consequently, an alternative loss function must be introduced that explicitly allows for speed inference.

Using this measurement, we incorporate an additional term in the loss function to ensure that speed is observable. 
We define the pursuer’s path length to a single interception location as
\begin{equation}
   L^{(i)}(\vect{\theta}_P,\vect{x}_s^{(i)}) = \rr\left(\vect{x}_s^{(i)}(t_f^{(i)}),\vect{\theta}_P\right) + R.
\end{equation}
The predicted launch time of the pursuer for the \(i^\text{th}\) sacrificial trajectory is
\begin{equation}
\hat{t}_{\ell}^{(i)}(\vect{\theta}_P,\vect{x}_s^{(i)}) = t_f^{(i)} - \frac{L^{(i)}(\vect{\theta}_P,\vect{x}_s^{(i)})}{v_P}.
\end{equation}
The time loss for a single sacrificial trajectory is defined ae
\begin{equation}
\label{eq:time_loss_with_flatten}
\mathcal{L}_{\text{time}}^{(i)}(\vect{\theta}_P,\vect{x}_s^{(i)},t_{\ell}^{(i)}) 
= \mathrm{ReSq}\!\big(|\hat{t}_{\ell}^{(i)}(\vect{\theta}_P,\vect{x}_s^{(i)}) - t_{\ell}^{(i)}| - \delta\big).
\end{equation}
where \(\delta = \beta\sigma_t\) is a flattening parameter that accounts for uncertainty in the launch-time measurement.  This loss function measures the difference between the measured launch time and the estimated launch time (derived from the inferred speed), giving higher values when the difference exceeds the threshold $\delta$.

Finally, for all trajectories, the total time loss is
\begin{equation}
\label{eq:total_time_loss}
\mathcal{L}_{\text{tot},\text{time}}(\vect{\theta}_P,\mathcal{X},\mathcal{T}_{\ell})
=\sum_{\vect{x}_s^{(i)}\in \mathcal{X},\, t_{\ell}^{(i)}\in \mathcal{T}_{\ell}}
\mathcal{L}_{\text{time}}^{(i)}(\vect{\theta}_P,\vect{x}_s^{(i)},t_{\ell}^{(i)}),
\end{equation}
where \(\mathcal{T}_{\ell} = \{t_{\ell}^{(1)}, \dots, t_{\ell}^{(N_s)}\}\) denotes the set of launch-time measurements.
To learn all the pursuer parameters, we combine the time loss with the previous loss from Equation~\eqref{eq:rs_total_loss}:
\begin{equation}
\label{eq:ez_total_loss}
\mathcal{L}_{\text{tot},\text{EZ}}(\vect{\theta}_P,\mathcal{X},\mathcal{Z},\mathcal{T}_{\ell})
= \mathcal{L}_{\text{tot,RR}}(\vect{\theta}_P,\mathcal{X},\mathcal{Z})
+ \mathcal{L}_{\text{tot},\text{time}}(\vect{\theta}_P,\mathcal{X},\mathcal{T}_{\ell}).
\end{equation}

\subsection{Loss Function: Interior Interception}\label{sec:loss_function_relaxed}
This section presents the loss function under the interior-interception assumption (\assump{A1-relaxed}), in which interception is not restricted to the RR boundary but may occur anywhere within the RR.

\subsubsection{Known Pursuer Speed (Cases~1 and~2)}
Under the interior-interception assumption (\assump{A1-relaxed}), the loss function is designed to satisfy the following properties:
\begin{enumerate}
    \item The interception point lies strictly inside the RR.
    \item A trajectory corresponding to a non-intercepted agent remains outside the RR at all times.
\end{enumerate}

In this case, there is no trajectory loss for intercepted outcomes, since the exact point at which the sacrificial agent entered the RR is unknown. The only available information is that, if the sacrificial agent survived, its trajectory never entered the RR. This leads to the trajectory loss function
\begin{equation}\label{eq:interior_traj_loss}
\mathcal{L}_{\text{traj,RR}}^{(i)}\left(\vect{\theta}_P,\vect{x}_s^{(i)}(t)\right)
=
\begin{cases}
\displaystyle \max_{t\in\vect{t}}\!\Big[\mathrm{ReSq}\!\big(-\rr(\vect{x}_s^{(i)}(t),\vect{\theta}_P)-\varepsilon\big)\Big], & z^{(i)}=0,\\[8pt]
0, & z^{(i)}=1,
\end{cases}
\end{equation}
which penalizes only surviving trajectories ($z^{(i)}=0$). Since $\rr(\vect{x},\vect{\theta}_P)$ is positive outside the RR and negative inside, the quantity
$\left(-\rr(\vect{x}_s^{(i)}(t),\vect{\theta}_P)-\varepsilon\right)$
becomes positive precisely when the trajectory enters the RR by more than the $\varepsilon$ margin. In this case, the loss becomes nonzero, indicating an inconsistency between the model prediction and the observed survival. If the trajectory remains outside the RR for all $t$, the argument of the $\mathrm{ReSq}(\cdot)$ is non-positive everywhere and the loss evaluates to zero. No trajectory penalty is applied for intercepted agents ($z^{(i)}=1$), since the entry point into the RR is unobserved.

Unlike the boundary condition in the \assump{A1-strict} case, where the interception point is required to lie exactly on the RR boundary, here we only require that it lie within the RR. This results in the boundary loss
\begin{equation}\label{eq:RR_bound_loss_interior}
\mathcal{L}_{\text{bound,RR}}^{(i)}(\vect{\theta}_P,\vect{x}_s^{(i)}(t),z^{(i)})
=
\begin{cases}
0, & z^{(i)}=0,\\[3pt]
\mathrm{ReSq}\!\big(\rr(\vect{x}_s^{(i)}(t_f),\vect{\theta}_P)-\varepsilon\big), & z^{(i)}=1,
\end{cases}
\end{equation}
which penalizes positive RR values, ensuring that the predicted interception point for an intercepted trajectory lies strictly within the RR of the candidate model.

As before, the boundary and trajectory losses are combined to form the total loss for each sacrificial agent,
\begin{equation}
\mathcal{L}_{\text{comb,RR}}^{(i)}(\vect{\theta}_P,\vect{x}_s^{(i)}(t),z^{(i)})
=
\mathcal{L}_{\text{bound,RR}}^{(i)}(\vect{\theta}_P,\vect{x}_s^{(i)}(t),z^{(i)})
+\mathcal{L}_{\text{traj,RR}}^{(i)}(\vect{\theta}_P,\vect{x}_s^{(i)}(t),z^{(i)}).
\end{equation}
For multiple trajectories (sacrificial agents), the total loss is obtained by summing across all agents:
\begin{equation}\label{eq:total_learning_loss_int}
\mathcal{L}_{\text{tot,RR}}(\vect{\theta}_P,\mathcal{X},\mathcal{Z})
=
\sum_{\vect{x}_s^{(i)}\in\mathcal{X},\,z^{(i)}\in\mathcal{Z}}
\mathcal{L}_{\text{comb,RR}}^{(i)}(\vect{\theta}_P,\vect{x}_s^{(i)}(t),z^{(i)}),
\end{equation}
where $\mathcal{X}$ and $\mathcal{Z}$ denote the sets of all sacrificial trajectories and corresponding outcomes, respectively. This relaxed-loss formulation ensures that intercepted trajectories are modeled as interior points of the RR while preserving the non-intercepted constraints, enabling parameter learning under partial or ambiguous interception information.

\subsubsection{Unknown Pursuer Speed (Cases 3)}
To learn the speed with the interior interception assumption, we use the same time loss term (Equation~\eqref{eq:total_time_loss}. We add this term to the total learning loss for Cases 1 and 2 (Equation~\eqref{eq:total_learning_loss_int};
\begin{equation}\label{eq:ez_total_loss_int}
\mathcal{L}_{\text{tot},\text{EZ}}(\vect{\theta}_P,\mathcal{X},\mathcal{Z},\mathcal{T}_{\ell})
= \mathcal{L}_{\text{tot,RR}}(\vect{\theta}_P,\mathcal{X},\mathcal{Z})
+ \mathcal{L}_{\text{tot},\text{time}}(\vect{\theta}_P,\mathcal{X},\mathcal{T}_{\ell}).
\end{equation}

\subsection{Parameter Estimation Via Nonlinear Optimization}\label{sec:optimization_strict}
To estimate the pursuer parameter vector $\vect{\theta}_P$, we minimize the total loss defined in 
Equation~\eqref{eq:rs_total_loss} for Cases~1 and~2, and 
Equation~\eqref{eq:ez_total_loss} for Case~3 under the boundary interception assumption. 
When using the interior interception assumption, we instead minimize 
Equation~\eqref{eq:total_learning_loss_int} for Cases~1 and~2, and 
Equation~\eqref{eq:ez_total_loss_int} for Case~3. 
This optimization problem is solved using IPOPT, a gradient-based interior-point optimizer~\cite{ipopt}. 
Lower and upper bounds are defined for each parameter in $\vect{\theta}_P$ to restrict the solution to physically plausible values. 
Let $\vect{\theta}_P^{\mathrm{min}}$ and $\vect{\theta}_P^{\mathrm{max}}$ denote the lower and upper bounds, respectively. 
The resulting optimization problem is
\begin{equation}
\label{eq:opt_problem}
\vect{\theta}_P^\star = 
\operatornamewithlimits{argmin}_{\vect{\theta}_P \in [\vect{\theta}_P^{\mathrm{min}},\, \vect{\theta}_P^{\mathrm{max}}]}
\mathcal{L}_{\mathrm{tot}}(\vect{\theta}_P, \mathcal{X}, \mathcal{Z}),
\end{equation}
where $\mathcal{L}_{\mathrm{tot}}(\vect{\theta}_P, \mathcal{X}, \mathcal{Z})$ denotes the total loss function for the corresponding case and interception assumption.
We compute the gradient of $\mathcal{L}_{\mathrm{tot}}$ with respect to $\vect{\theta}_P$ using JAX’s automatic differentiation framework~\cite{jax2018github}, which enables efficient and scalable gradient evaluation within the IPOPT solver.

\subsection{Multi-start Estimation Strategy}\label{sec:particle_strict}
% To address the nonconvexity of the estimation problem and the potential for multiple local minima, we adopt a multi-start inference scheme rather than relying on a single optimizer initialization. 
% To mitigate the nonconvexity of the estimation problem and the risk of converging to local minima, we employ a multi-start inference scheme instead of relying on a single optimizer initialization.
% This approach maintains a population of candidate parameter estimates that are repeatedly refined as new data become available. By doing so, the method avoids premature convergence to a single mode and improves robustness in multimodal landscapes.
To mitigate the nonconvexity of the estimation problem and the resulting ambiguity in the inferred parameters, we employ a multi-start inference scheme rather than relying on a single optimizer initialization. In practice, the loss landscape induced by the hinged-based constraints contains large flat regions of zero loss: many distinct parameter vectors produce RR geometries that are fully consistent with the observed data and therefore achieve identical loss values of zero.

As a result, different optimizer initializations often converge to different parameter estimates, all of which satisfy the imposed consistency conditions but may correspond to substantially different RR geometries and yield different predictions for new sacrificial trajectories. Single-start optimization is therefore insufficient, as it returns only one such solution without representing the inherent ambiguity supported by the available data.

The proposed approach maintains a population of candidate parameter estimates that are refined as new data become available. By preserving multiple hypotheses in parallel, the method captures this ambiguity and enables solutions to be progressively refined as additional measurements are incorporated.

Formally, let $N_p$ denote the number of optimizer initializations. We find the $N_p$ starting values using Latin Hypercube Sampling (LHS) to uniformly cover the prior domain $[\vect{\theta}_P^{\mathrm{min}}, \vect{\theta}_P^{\mathrm{max}}]$. Each sampled initialization $\vect{\theta}_P^{(j)}$ for $j = 1, \dots, N_p$ is used as the starting point for a separate IPOPT optimization, producing a set of locally optimized solutions:
\begin{equation}
\mathcal{T}=\left\{ \vect{\theta}_P^{(j)*} \right\}_{j=1}^{N_p},  \quad \vect{\theta}_P^{(j)*} = \arg\min_{\vect{\theta}_P} \mathcal{L}_{\mathrm{tot}}(\vect{\theta}_P,\mathcal{X},\mathcal{Z}). \label{eq:particle_solutions} 
\end{equation}

After this batch of optimizations, we select a new sacrificial agent path (described in Section~\ref{sec:sacrificial_path}) and observe the result of its trajectory. The resulting trajectory and interception outcome are added to the dataset. The full process—optimization, agent deployment, and data update—is then repeated with the previously optimized pursuer parameter value serving as the starting points for the next round. This iterative scheme allows the model to progressively refine the estimate of $\vect{\theta}_P$ as more information becomes available. Sacrificial agents are dispatched until the standard deviation of the estimates is below a threshold. 
% Care must be taken when computing the mean and standard deviation of the heading, using the circular mean and circular standard deviation.
% Because heading is an angular quantity, we compute summary statistics using standard circular (trigonometric-moment) definitions. Given angle samples $\{\psi_i\}_{i=1}^N$, we form
% $C=\frac{1}{N}\sum_{i=1}^N \cos\psi_i$ and $S=\frac{1}{N}\sum_{i=1}^N \sin\psi_i$, and compute the circular mean
% \begin{equation}
% \mu_{\psi}=\mathrm{atan2}(S,C),
% \end{equation}
% wrapped to $(-\pi,\pi]$. The mean resultant length $R=\sqrt{C^2+S^2}$ quantifies angular concentration; we report the circular variance $v = 1-R$ and the circular standard deviation
% \begin{equation}
% \sigma_\psi=\sqrt{-2\ln R},
% \end{equation}
% with a maximum-entropy fallback when $R$ is numerically near zero.

Our full estimation algorithm proceeds as follows:
\begin{enumerate}
    \item \textbf{Initialize.} Draw $N_p$ initializations $\{\vect{\theta}_P^{(j)}\}_{j=1}^{N_p}$ via Latin Hypercube Sampling within the prior bounds, and set $\mathcal{X}\leftarrow\emptyset$ and $\mathcal{Z}\leftarrow\emptyset$. Initialize $\sigma_{\vect{\theta}_P}\gets\infty$.

    \item \textbf{Repeat until converged (elementwise).} While $\operatorname{any}\!\left(\sigma_{\vect{\theta}_P} > \sigma_{\text{thresh}}\right)$, i.e., while any component of the estimated standard deviation vector exceeds its corresponding threshold:
    \begin{enumerate}[label=\alph*)]
        \item \textbf{Plan \& deploy a sacrificial agent.} Select the next trajectory (Section~\ref{sec:sacrificial_path}), execute it, and collect the observed path $\vect{x}_s^{(k)}(t)$ and outcome $z^{(k)}$.

        \item \textbf{Update the dataset.} Append $\vect{x}_s^{(k)}(t)$ to $\mathcal{X}$ and $z^{(k)}$ to $\mathcal{Z}$.

        \item \textbf{Optimize and filter.} For $j=1,\dots,N_p$, solve
        \[
        \vect{\theta}_P^{(j)*} \leftarrow \arg\min_{\vect{\theta}}\ \mathcal{L}(\vect{\theta};\mathcal{X},\mathcal{Z})
        \]
        (Eq.~\eqref{eq:opt_problem}), and discard any candidate with $\mathcal{L}(\vect{\theta}_P^{(j)*};\mathcal{X},\mathcal{Z})>\epsilon_{\mathcal{L}}$.

        \item \textbf{Compute summary statistics.} Compute $\mu_{\vect{\theta}_P}$ and $\sigma_{\vect{\theta}_P}$ from the retained (feasible) candidates.

        \item \textbf{Resample starting points.} Form the next set of $N_p$ initializations by keeping all retained candidates and randomly resampling (with replacement) and jittering them until $N_p$ starting points are obtained.
    \end{enumerate}

    \item \textbf{Terminate.} When all components satisfy their thresholds, return the final retained candidates as the estimate of $\mathcal{T}=\{\vect{\theta}_P^{(j)*}\}$.
\end{enumerate}

\begin{algorithm}[htbp]
\caption{Multi-start Inference of Pursuer Parameters}
\label{alg:multistart_inference}
\begin{algorithmic}[1]
\Require Number of initializations $N_p$, prior bounds on $\vect{\theta}_P$, tolerances $\sigma_\textit{thresh}$ and loss threshold $\epsilon_{\mathcal{L}}$
\State Initialize candidate set $\{ \vect{\theta}_P^{(j)} \}_{j=1}^{N_p}$ via Latin Hypercube Sampling over prior bounds
\State Initialize measurement collections $\mathcal{X} \gets \emptyset$, $\mathcal{Z} \gets \emptyset$
\State Initialize $\sigma_{\vect{\theta}_P} \gets \infty$; trial index $k \gets 0$
\While{$\operatorname{any}(\sigma_{\vect{\theta}_P} > \sigma_\textit{thresh})$}
    \State $k \gets k+1$
    \State Dispatch $k^\text{th}$ sacrificial agent; observe $\vect{x}_s^{(k)}(t)$ and $z^{(k)}$
    \State $\mathcal{X} \gets \mathcal{X} \cup \{ \vect{x}_s^{(k)}(t) \}$, \quad $\mathcal{Z} \gets \mathcal{Z} \cup \{ z^{(k)} \}$
    \For{$j = 1$ to $N_p$}
        \State $\vect{\theta}_P^{(j)*} \leftarrow \arg\min_{\vect{\theta}} \mathcal{L}(\vect{\theta}; \mathcal{X}, \mathcal{Z})$ \Comment{Eq.~\eqref{eq:opt_problem}}
    \EndFor
    \State Retain candidates with $\mathcal{L}(\vect{\theta}_P^{(j)*};\mathcal{X},\mathcal{Z}) \le \epsilon_{\mathcal{L}}$
    \State Compute $\mu_{\vect{\theta}_P}$ and $\sigma_{\vect{\theta}_P}$ from the retained candidates
    \State Resample (with replacement) from retained candidates and apply small random perturbations (jitter) until $N_p$ candidates are obtained
\EndWhile
\State \Return Retained candidate set $\mathcal{T}=\{\vect{\theta}_P^{(j)*}\}$
\end{algorithmic}
\end{algorithm}

\begin{figure}[H]
    \centering
\includegraphics[width=.6\linewidth,trim={0.0cm 0.cm .0cm .0cm},clip]{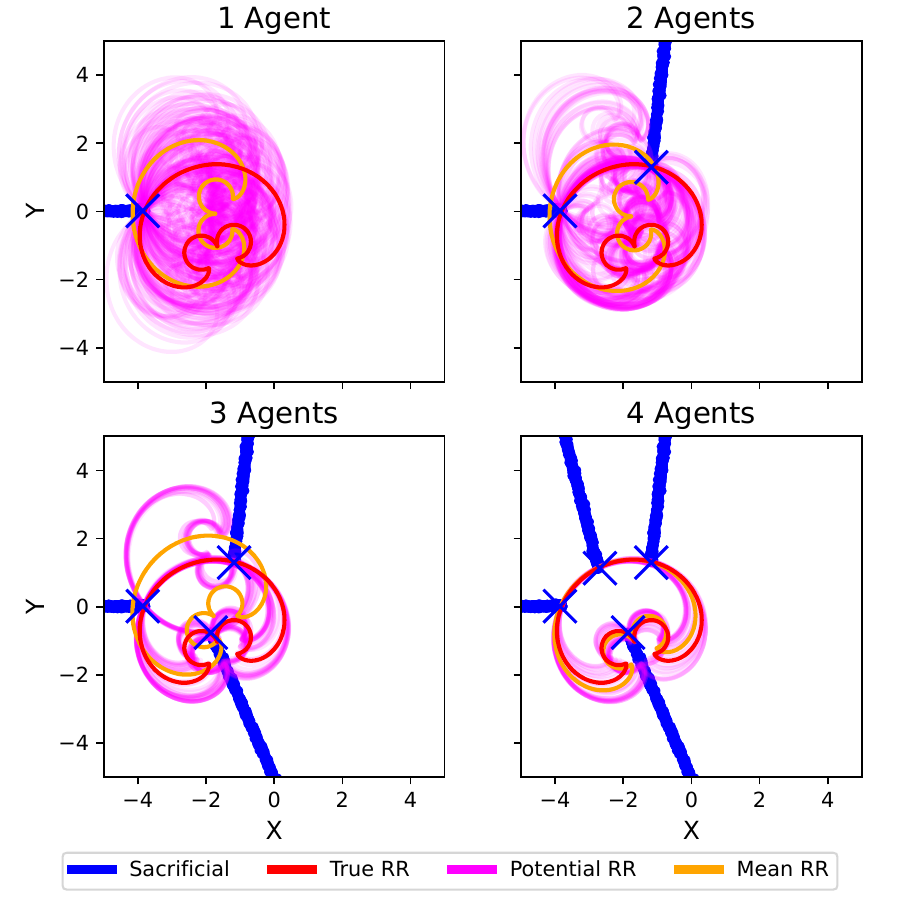}
\caption{Sacrificial trajectories (blue), along with the true (red), potential (magenta), and learned mean (orange) pursuer reachable regions.}\label{fig:examples_trajectories}
\end{figure}

\begin{figure}[ht]
    \centering
\includegraphics[width=.8\linewidth,trim={0.0cm 0.cm .0cm .0cm},clip]{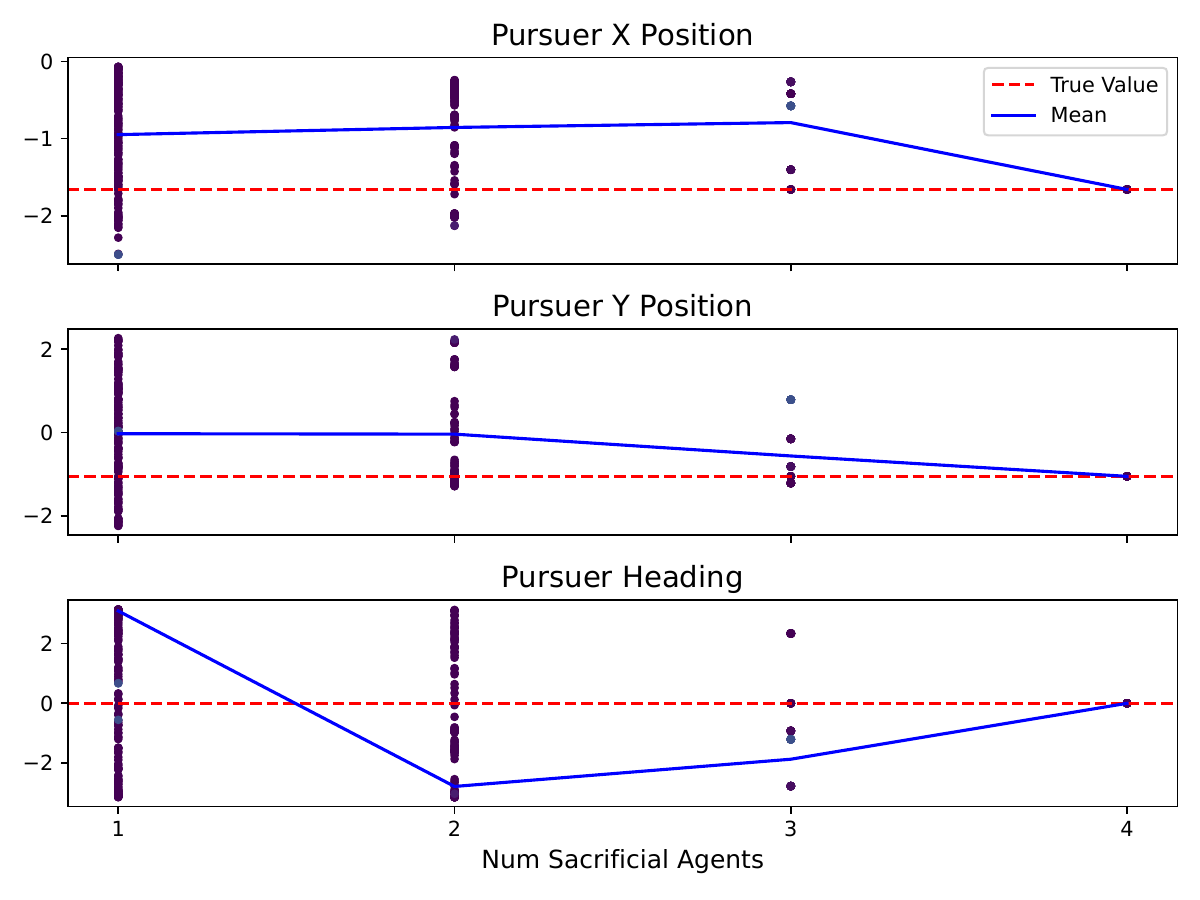}
\caption{Learned parameter values (points) and the mean learned parameter value (blue), along with the true parameter value (red).}\label{fig:sample_spread}
\end{figure}

To illustrate the mechanics of the learning procedure (rather than to evaluate performance), we include a qualitative example trial from Case~2B (unknown $x_P,y_P,\psi_P,a,R$ with noisy trajectory measurements). Figures~\ref{fig:examples_trajectories} and~\ref{fig:sample_spread} visualize how the feasible RR family contracts as additional sacrificial trials are incorporated.

Figure~\ref{fig:examples_trajectories} illustrates representative sacrificial-agent trials for increasing numbers of agents (one through four, left-to-right and top-to-bottom). The executed sacrificial trajectories are shown in blue. Also shown are the true RR (red), the family of RRs induced by the feasible parameter set $\mathcal{T}$ (magenta), and the RR generated by the mean estimate $\mu_{\vect{\theta}_P}$ (orange). The envelope of feasible RRs contains the true RR, and as additional sacrificial trials are incorporated this envelope contracts toward the true RR. With a single agent, many distinct pursuer parameter vectors remain feasible and produce a wide range of admissible RRs; as the number of agents increases, inconsistent parameter hypotheses are eliminated and the feasible RR family collapses around the true RR.

Similarly, Figure~\ref{fig:sample_spread} shows the true (red) and mean (blue) of the set of estimated pursuer parameter vectors. Each optimized parameter vector is also plotted (black points), showing how the set of potential pursuer parameter vectors collapses to a single value. 

The resulting set of feasible pursuer parameter vectors not only supports parameter inference but also forms the basis for downstream planning; in Section~\ref{sec:safe_path} we show how these learned engagement regions can be incorporated into real-time trajectory optimization.

\section{Sacrificial Agent Trajectory Selection}\label{sec:sacrificial_path}
To maximize the efficiency of sacrificial agents, trajectories should be selected that will yield the greatest information learned about the pursuer’s RR (or EZ). We consider two complementary strategies: (i) a geometry-based approach that plans trajectories to maximize the dispersion of possible intercept locations around the RR shown in Section~\ref{sec:boundary_sac_planning}, and (ii) an information-based approach that uses the curvature of the Gauss--Newton information matrix of the loss shown in Section~\ref{sec:int_sac_planning}.

\subsection{Boundary Interception Trajectory Selection}\label{sec:boundary_sac_planning}
For the boundary assumption (\assump{A1-strict}), the most informative trajectories are those that result in interceptions maximally spread around the boundary of the pursuer’s reachable region (RR). However, the true boundary is unknown to the sacrificial agents. The only available information is the current set of pursuer parameters that satisfy the loss function, denoted by $\mathcal{T}$. The goal is therefore to identify sacrificial trajectories whose interception locations are distributed as widely as possible along this inferred boundary. To achieve this, we maximize the expected distance between the predicted future interception location and all previously observed interception locations.

For a candidate sacrificial trajectory $\vect{x}_s$, each potential parameter vector $\vect{\theta}_P^{(j)}\in\mathcal{T}$ that predicts an interception induces a predicted interception location, obtained by evaluating the RR function along $\vect{x}_s$ and identifying the first point at which $\rr$ changes sign from positive to negative. Collecting these points yields
\begin{equation}
\{\vect{x}_{\text{hit}}^{(j)}(\vect{x}_s)\}_{j=1}^{N_{p,\text{hit}}(\vect{x}_s)},
\end{equation}
where $N_{p,\text{hit}}(\vect{x}_s)$ is the number of parameter vectors in $\mathcal{T}$ that predict an interception for $\vect{x}_s$.

We summarize the predicted interception distribution for $\vect{x}_s$ by its centroid,
\begin{equation}
\vect{x}_{\text{hit}}(\mathcal{T},\vect{x}_s)=\frac{1}{N_{p,\text{hit}}(\vect{x}_s)}
\sum_{j=1}^{N_{p,\text{hit}}(\vect{x}_s)} \vect{x}_{\text{hit}}^{(j)}(\vect{x}_s).
\end{equation}

The next sacrificial trajectory is then selected by optimizing an exploration objective over candidate trajectories; specifically, we choose $\vect{x}_s^{(i+1)}$ to maximize the minimum distance between $\vect{x}_{\text{hit}}(\mathcal{T},\vect{x}_s)$ and the set of previously observed interception locations:
\begin{equation}
    \operatorname*{argmax}_{\vect{x}_s^{(i+1)}} \; 
    \min_{i} \;
    \big\| \vect{x}_{\text{hit}} - \vect{x}_{s}^{(i)}(t_f^{(i)}) \big\|,
\end{equation}
for all trajectories $\vect{x}_{s}^{(i)}$ that resulted in an interception.

Rather than optimizing over the entire trajectory space, we parameterize the sacrificial path with a constrained starting point and heading. Specifically, the starting point is restricted to lie on a circle of radius $r_s$ centered at the current estimated pursuer position $\mu_{\vect{\theta}_P}$. A candidate trajectory is therefore completely defined by two parameters: the angle $\alpha^{(i+1)}$ specifying the starting point on the circle, and the heading $\psi_s^{(i+1)}$. In practice, sacrificial agents would first travel to this starting point and then execute the prescribed straight-line trajectory; however, in this work, we assume that the agent begins directly at the trajectory’s starting position.

The objective function defined above is both non-differentiable and non-convex. Consequently, we employ an exhaustive grid-based search over the two parameters that define each sacrificial trajectory. Leveraging GPU-based parallelization, this search can be executed efficiently and has produced empirically near-global solutions in all tested cases. For comparison, we also applied gradient-based optimization to the utility function. Using standard single-start initializations, the optimizer consistently converged to lower-utility solutions. We additionally initialized the gradient-based optimizer at the best grid-search point and reran the optimization; this yielded only marginal improvement, which further motivated the direct grid-based approach used here.
% For comparison, we also implemented gradient-based optimization of the utility function initialized naively, which consistently converged to inferior local optima. Using the grid search to provide initial guesses offered only marginal improvement, further motivating the choice of a direct grid-based approach.

\subsection{Interior Interception Trajectory Selection}\label{sec:int_sac_planning}

Under the interior-interception assumption \assump{A1-relaxed}, the most informative sacrificial trajectories are those that skim just outside the RR. Each sacrificial trajectory can be interpreted as an experiment probing the unknown pursuer parameters. Trajectories that remain far outside the RR are unlikely to be intercepted and thus provide little discriminatory information, while trajectories that penetrate deeply into the RR are almost certainly intercepted and therefore also offer limited information for distinguishing between parameter vectors. The most informative trajectories lie near the RR boundary, where small variations in the pursuer parameters strongly influence the interception outcome. The proposed BED formulation formalizes this intuition by selecting trajectories that maximize the expected information gain about the pursuer parameters, accounting for both possible outcomes (interception and survival) and their predicted likelihoods.

Because the learning loss is based on squared hinge penalties, it exhibits locally quadratic structure near active constraints. This motivates the use of a Gauss--Newton-style approximation to the loss curvature with respect to the pursuer parameters. We adopt this curvature as an information surrogate and apply a D-optimality criterion to select sacrificial trajectories that are maximally informative for parameter estimation.

Given a candidate trajectory $\vect{x}_s^{(i+1)}$, we seek to evaluate its expected information gain. For each potential pursuer vector $\vect{\theta}_P^{(j)} \in \mathcal{T}$, we evaluate the RR along the trajectory $\vect{x}_s^{(i+1)}$. If any point along the trajectory lies inside the RR of the parameter vector, we declare that trajectory intercepted under that potential pursuer. This produces a binary outcome
\begin{equation}
    z^{(j)} \in \{0,1\},
\end{equation}
indicating whether interception occurs.

We also approximate the expected interception location under each potential parameter vector. Unlike the boundary-interception case, here interception is assumed to occur in the interior of the RR. We assume that the sacrificial agent is equally likely to be intercepted at any point along the portion of its trajectory lying inside the RR. The predicted interception location for parameter vector $\vect{\theta}_P^{(j)}$ is then defined as the average of the trajectory points with negative RR values, yielding
\begin{equation}
    \{\vect{x}_{\text{hit}}^{(j)}\}_{j=1}^{N_{P}}.
\end{equation}

We compute the empirical interception probability across the set of all inferred parameter vectors as
\begin{equation}
    \hat p = \frac{1}{N_p} \sum_{j=1}^{N_p} z^{(j)},
\end{equation}
and the average predicted interception location as
\begin{equation}
    \vect{x}_{\text{hit}}(\mathcal{T},\vect{x}_s^{(i+1)}) =
    \frac{\sum_{j=1}^{N_p} z^{(j)}\,\vect{x}_{\text{hit}}^{(j)}}{\sum_{j=1}^{N_p} z^{(j)}}.
\end{equation}

To compute the expected information gain we construct two hypothetical trajectories: an intercepted trajectory $\vect{x}_{s,\text{hit}}^{(i+1)}(t)$ that terminates at the average interception location $\vect{x}_{\text{hit}}(\mathcal{T},\vect{x}_s^{(i+1)})$, and a non-intercepted trajectory $\vect{x}_{s,\text{miss}}^{(i+1)}(t)$ that continues unimpeded.

The information contribution for the intercepted case corresponds to the interception-consistency loss in \eqref{eq:RR_bound_loss_interior}. For the intercepted trajectory, we evaluate the RR at the average hit location and define the informativeness weight
\begin{equation}
    \omega_{\text{hit}} = \mathrm{ReLU}\!\bigl(\rr(\vect{x}_{\text{hit}};\vect{\theta}_P^{(j)}) - \varepsilon\bigr),
\end{equation}
which quantifies the degree of inconsistency between the parameter vector prediction and the observed interception outcome. When the predicted hit location lies strictly within the RR, $\omega_{\text{hit}} = 0$ and the observation is consistent with the parameters, contributing no information under the adopted loss. When the predicted hit lies outside the RR, $\omega_{\text{hit}}$ becomes positive and increases with the degree of violation.

The gradient of the RR with respect to the pursuer parameters at the hit location is
\begin{equation}
g_{\text{hit}} = \nabla_{\vect{\theta}_P}\,\rr(\vect{x}_{\text{hit}};\vect{\theta}_P^{(j)}).
\end{equation}
Under the squared-hinge loss, the gradient of the corresponding loss term is proportional to $\omega_{\text{hit}} g_{\text{hit}}$. We therefore define a loss-induced information surrogate via the outer product of this gradient, yielding a weighted Gauss--Newton approximation to the curvature:
\begin{equation}
    I_{\text{hit}}(\vect{x}_s^{(i+1)},\vect{\theta}_P^{(j)}) 
    = \omega_{\text{hit}}^2\, g_{\text{hit}} g_{\text{hit}}^\top.
\end{equation}

For the missed trajectory $\vect{x}_{s,\text{miss}}^{(i+1)}(t)$, we compute the information contribution corresponding to the trajectory loss in \eqref{eq:interior_traj_loss}. We discretize the trajectory at time points $t_i \in \vect{t} = \{t_1,\dots,t_{N_t}\}$ and evaluate the hinge violation along the path:
\begin{equation}
    \omega_i = \mathrm{ReLU}\!\bigl(-\,\rr(\vect{x}_{s,\text{miss}}^{(i+1)}(t_i),\vect{\theta}_P^{(j)}) - \varepsilon\bigr).
\end{equation}
The scalar weight $\omega_i$ quantifies the degree of parameter--data inconsistency for the survival outcome, with larger values indicating stronger violation of the requirement that a surviving trajectory remain outside the RR.
We then select the maximally violating point along the trajectory,
\begin{equation}
    i^\star = \arg\max_i \; \omega_i,
\end{equation}
and the corresponding gradient
\begin{equation}
    g_{\text{miss}} = \nabla_{\vect{\theta}_P}\,\rr(\vect{x}_{s,\text{miss}}^{(i+1)}(t_{i^\star}),\vect{\theta}_P^{(j)}).
\end{equation}
The corresponding Gauss--Newton curvature surrogate is
\begin{equation}
    I_{\text{miss}}(\vect{x}_s^{(i+1)},\vect{\theta}_P^{(j)})
    = \omega_{i^\star}^2\, g_{\text{miss}} g_{\text{miss}}^\top.
\end{equation}

The expected information increment for trajectory $\vect{x}_s^{(i+1)}$ under parameters $\vect{\theta}_P^{(j)}$ is
\begin{equation}
    \Delta I(\vect{x}_s^{(i+1)},\vect{\theta}_P^{(j)}) =
    \hat p\,I_{\text{hit}}(\vect{x}_s^{(i+1)},\vect{\theta}_P^{(j)})
    + (1-\hat p)\,I_{\text{miss}}(\vect{x}_s^{(i+1)},\vect{\theta}_P^{(j)}).
\end{equation}
Averaging across the ensemble yields the expected information increment used for trajectory selection:
\begin{equation}
    \Delta I_{\text{avg}}(\vect{x}_s^{(i+1)},\mathcal{T}) =
    \frac{1}{N_p}\sum_{j=1}^{N_p}\Delta I(\vect{x}_s^{(i+1)},\vect{\theta}_P^{(j)}).
\end{equation}

For past (realized) trajectories, the outcome is known. The information increment for a single trajectory $\vect{x}_s^{(i)}$ and parameter vector $\vect{\theta}_P^{(j)}$ is therefore
\begin{equation}
    \Delta I_{\text{past}}(\vect{x}_s^{(i)},z^{(i)},\vect{\theta}_P^{(j)}) =
    z^{(i)}\,I_{\text{hit}}(\vect{x}_s^{(i)},\vect{\theta}_P^{(j)})
    + (1-z^{(i)})\,I_{\text{miss}}(\vect{x}_s^{(i)},\vect{\theta}_P^{(j)}).
\end{equation}
Here, the true interception location is used rather than the predicted location. Averaging over the set of all parameter vectors yields
\begin{equation}
    \Delta I_{\text{avg,past}}(\vect{x}_s^{(i)},z^{(i)},\mathcal{T}) =
    \frac{1}{N_p}\sum_{j=1}^{N_p}\Delta I_{\text{past}}(\vect{x}_s^{(i)},z^{(i)},\vect{\theta}_P^{(j)}),
\end{equation}
and accumulating over all past trajectories gives
\begin{equation}
     I_{\text{past}}(\mathcal{X},\mathcal{Z},\mathcal{T}) =
     \sum_{i=1}^{N_s} \Delta I_{\text{avg,past}}(\vect{x}_s^{(i)},z^{(i)},\mathcal{T}).
\end{equation}

We now have both the accumulated curvature surrogate from past trajectories and the expected change induced by executing the candidate trajectory $\vect{x}_s^{(i+1)}$. We compute the D-optimality score
\begin{equation}\label{eq:d_score}
    D_{\text{gain}} =
    \log\det\!\left(I_{\text{past}}+ \Delta I_\text{avg}(\vect{x}_s^{(i+1)})\right) 
    - \log\det\!\left(I_{\text{past}}\right).
\end{equation}
The next sacrificial trajectory is selected by maximizing this D-optimality score, yielding the trajectory that maximally increases the curvature surrogate and thus steepens the loss landscape under the hinge-based learning objective.

As in the boundary-interception planner, we restrict the search space by constraining the trajectory start location to lie on a circle. Each candidate trajectory is parameterized by two variables: the angular start position $\alpha^{(i+1)}$ and the initial heading $\psi_s^{(i+1)}$. The trajectory defined by these parameters that maximizes the D-score is selected for execution.

While this section focused on selecting sacrificial trajectories to refine pursuer estimates, the resulting learned feasible-parameter set can also be used for motion planning of high-priority agents requiring safe paths. The following section demonstrates how the learned engagement zones are used to compute safe, time-optimal trajectories.

\section{Path Planning with Learned Feasible Engagement Zones}\label{sec:safe_path}
In this section, we demonstrate how the learned set of feasible pursuer parameter vectors $\mathcal{T}$, can be directly leveraged for real-time path planning. By constructing EZs for each parameter vector in~$\mathcal{T}$,
we form a threat map that captures all pursuer parameters consistent with the observations. The proposed planning algorithm then computes a trajectory that safely navigates around these EZs, ensuring that the evader remains outside every learned, feasible interception region. This integration connects parameter learning with trajectory planning and enables the
generation of shorter paths that satisfy safety constraints.
`

We parameterize the evader’s trajectory using a B-spline representation and employ a nonlinear optimizer to minimize the total path time, which is equivalent to the path length, since the evader travels at constant velocity. Kinematic feasibility is enforced through constraints on maximum turn rate and curvature to ensure the resulting trajectory is flyable. Additionally, the path is constrained to remain outside all EZs generated from the all pursuer parameter vectors in~$\mathcal{T}$, guaranteeing safety against every learned pursuer configuration.

We represent the evader trajectory with a B-spline curve due to its smoothness, differentiability, and local support, which together yield sparse Jacobians well-suited to gradient-based optimization. Let the control points be $\mathcal{C} = (\vect{c}_1, \vect{c}_2, \ldots, \vect{c}_{N_c})$,
and define the knot vector $\vect{t}_k = \big(t_0 - k\Delta_t, \ldots, t_0 - \Delta_t,\, t_0,\,
t_0 + \Delta_t, \ldots, t_f,\, t_f + \Delta_t, \ldots, t_f + k\Delta_t\big)$, where $t_0$ and $t_f$ are the start and end times, $k$ is the spline order, $N_k$ is the number of internal knots, and $\Delta_t = (t_f - t_0)/N_k$ is the uniform knot spacing. The position at time $t$ is a weighted sum of basis functions:
\begin{equation}
\label{eq:bspline}
\vect{p}(t) = \sum_{i=1}^{N_c} B_{i,k}(t)\,\vect{c}_i,
\end{equation}
with $\{B_{i,k}\}$ defined recursively via the Cox--de~Boor construction~\cite{cox1972numerical}.

We seek a time-optimal (equivalently, length-optimal under constant speed) trajectory
from an initial location $\vect{x}_0$ to a goal location $\vect{x}_f$.
The set of potential pursuer parameter vectors $\mathcal{T}$ induces EZ constraints
that require the trajectory to avoid capture.
Trajectory feasibility is enforced using a unicycle kinematic model with bounded turn
rate and curvature.
The resulting nonlinear optimization is
\begin{subequations}
\label{eq:optimization}
\begin{align}
(\mathcal{C}_{\mathrm{opt}},\, t_{f_{\mathrm{opt}}})
&= \arg\min_{\mathcal{C},\, t_f} \; t_f, \label{eq:opt_obj}\\[2pt]
\text{s.t.}\quad
&\vect{p}(0) = \vect{x}_0, \label{eq:opt_start}\\
&\vect{p}(t_f) = \vect{x}_f, \label{eq:opt_end}\\
% &P_{\ez}\!\big(\mu_{\vect{\theta}_P}, \Sigma_{\vect{\theta}_P}, \vect{\theta}_E(\vect{t}_s)\big) \le \epsilon, \label{eq:opt_ez}\\
&\ez(\vect{p}(\vect{t}_s),\psi(\vect{p}(\vect{t}_s)),\vect{\theta}_P^{(j)})\geq 0; \vect{\theta}_P^{(j)}\in \mathcal{T}, \label{eq:opt_ez}\\
&v(\vect{t}_s) = v_E, \label{eq:opt_speed}\\
&u_{\mathrm{lb}} \le u_E(\vect{t}_s) \le u_{\mathrm{ub}}, \label{eq:opt_turnrate}\\
&|\kappa_E(\vect{t}_s)| \le \kappa_{\mathrm{ub}}, \label{eq:opt_curvature}
\end{align}
\end{subequations}
where $\psi(\vect{p}(t)) =\mathrm{atan2}(\dot{p}_y(t),\, \dot{p}_x(t))$ is the heading of the trajectory, $\vect{t}_s = \{0,\Delta_t,2\Delta_t,\ldots,t_f\}$ are the discrete times to enforce the constraints at with $\Delta_t = t_f/N_t$. Constraints~\eqref{eq:opt_start}--\eqref{eq:opt_end} set the boundary conditions; \eqref{eq:opt_ez} encodes avoidance of all EZs generated by $\mathcal{T}$ (recall $\ez(\vect{x}_E, \psi_E, \vect{\theta}_P)\leq0$ means the evader is capturable); and \eqref{eq:opt_speed}--\eqref{eq:opt_curvature} ensure kinematic feasibility. Under the unicycle model,
\begin{equation}
v(t) = \|\dot{\vect{p}}(t)\|_2,\qquad
u(t) = \frac{\dot{\vect{p}}(t) \times \ddot{\vect{p}}(t)}{\|\dot{\vect{p}}(t)\|_2^2},\qquad
\kappa(t) = \frac{u(t)}{v(t)}.
\end{equation}

We solve~\eqref{eq:optimization} using the interior-point solver IPOPT~\cite{ipopt}. All required derivatives (objective and constraints) are obtained via automatic differentiation in JAX~\cite{jax2018github}, enabling efficient assembly of sparse Jacobians and Hessian approximations.

\section{Results}\label{sec:results}
In this section, we present the results of the pursuer-parameter learning algorithm. We conduct 500 Monte Carlo (MC) simulations using 500 distinct pursuer parameter vectors, each sampled uniformly from the range $[\vect{\theta}_P^\text{min}, \vect{\theta}_P^\text{max}]$. The same parameter vectors are used across all evaluated cases to ensure consistent comparison. 
For each run, we compute the absolute error between the true and estimated pursuer parameter values. In addition, we evaluate two geometric consistency metrics based on the resulting RRs. First, we report the ratio between the area of the true RR and the area of the union of all RRs generated from the feasible set of learned parameter vectors. Second, we compute the percentage of the true RR that is covered by this union. Both metrics are obtained by numerically integrating over the RR areas.

\subsection{Boundary Interception}
\begin{figure}[ht]
    \centering
\includegraphics[width=.95\linewidth,trim={0.0cm 0.cm .0cm .0cm},clip]{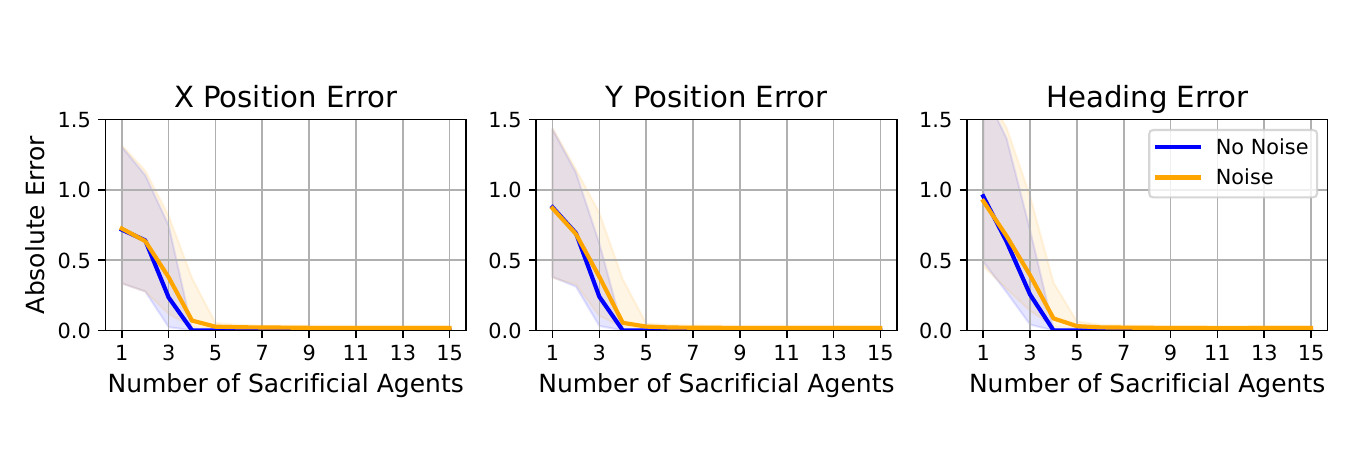}
\caption{Median absolute error (solid) and IQR (shaded) across 500 MC runs for Cases 1A and 1B under the boundary-interception assumption. Blue indicates no measurement noise, and orange indicates noisy measurements.}\label{fig:error_known_shape}
\end{figure}

We report the absolute error from the estimated mean value of the pursuer's parameters and the true value. For Case 1, only the location and orientation of the pursuer are learned. Figure~\ref{fig:error_known_shape} shows the median error for each learned parameter for both the noise free (1A, blue) and noisy (1B, orange) cases. In this and in subsequent plots, the gray and orange shaded regions represent the interquartile range (IQR) for the noise free and noisy cases, respectively.  As can be seen, it typically takes four sacrificial agents to learn the position and orientation of the pursuer in the noise free case, and five agents in the noisy case. 

\begin{figure}[ht]
    \centering
\includegraphics[width=.95\linewidth,trim={0.0cm 0.cm .0cm .0cm},clip]{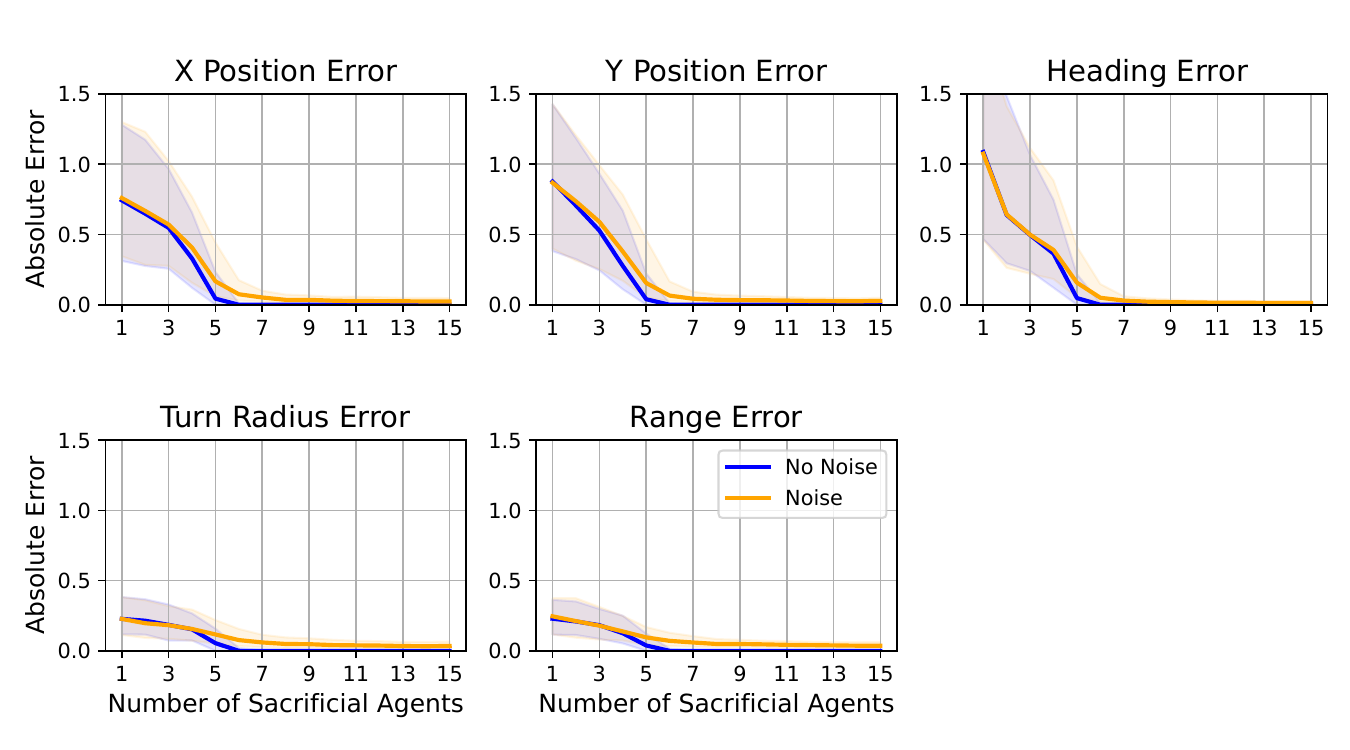}
\caption{Median absolute error (solid) and IQR (shaded) across 500 runs for Cases 2A and 2B under the boundary-interception assumption. Blue indicates no measurement noise, and orange indicates noisy measurements.}\label{fig:error_known_speed}
\end{figure}

Figure~\ref{fig:error_known_speed} shows results for Cases 2A (blue) and 2B (orange) under the boundary interception assumption. The plots report the median absolute error between the true and estimated pursuer parameters. In the noise-free case, roughly five sacrificial agents are sufficient to recover accurate parameter estimates. With noisy measurements, the error decreases after about seven agents, with reliable convergence achieved after approximately nine agents.

\begin{figure}[ht]
    \centering
\includegraphics[width=.95\linewidth,trim={0.0cm 0.cm .0cm .0cm},clip]{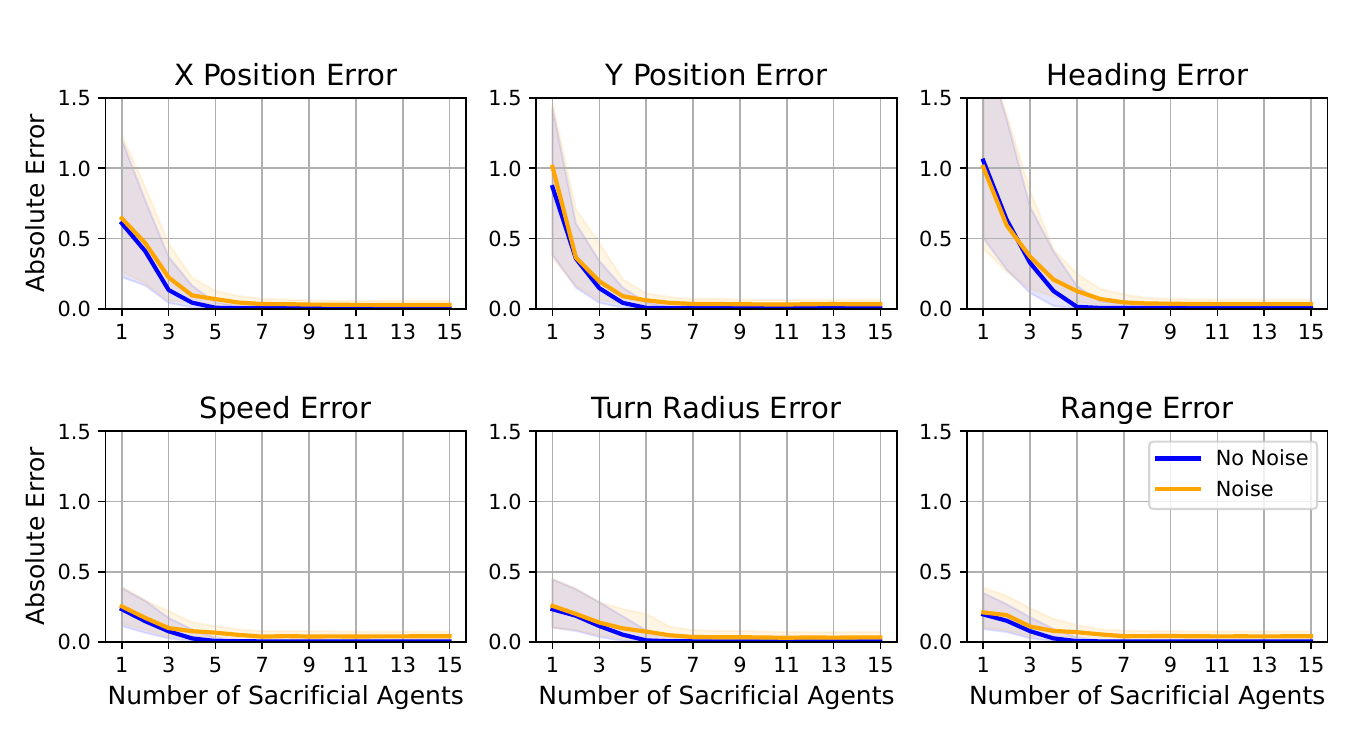}
\caption{Median absolute error (solid) and IQR (shaded) across 500 runs for Cases 3A and 3B under the boundary-interception assumption. Blue indicates no measurement noise, and orange indicates noisy measurements.}\label{fig:error_unknown_speed}
\end{figure}

Figure~\ref{fig:error_unknown_speed} shows Cases 3A (blue) and 3B (orange) under the boundary interception assumption. Plotted is the median absolute error between the true and estimated values of each pursuer parameter, with shaded regions indicating the interquartile range across 500 Monte Carlo trials. In the noise-free case, parameter errors converge after about five sacrificial agents. With noisy trajectories, the errors decrease more gradually, requiring approximately seven agents for reliable convergence across all parameters. Even though there is an extra unknown variable compared to Case 2, it takes around the same number of agents to converge because there is an extra measurement (launch time) in this case.

\begin{figure}[ht]
    \centering
\includegraphics[width=.95\linewidth,trim={0.0cm 0.cm .0cm .0cm},clip]{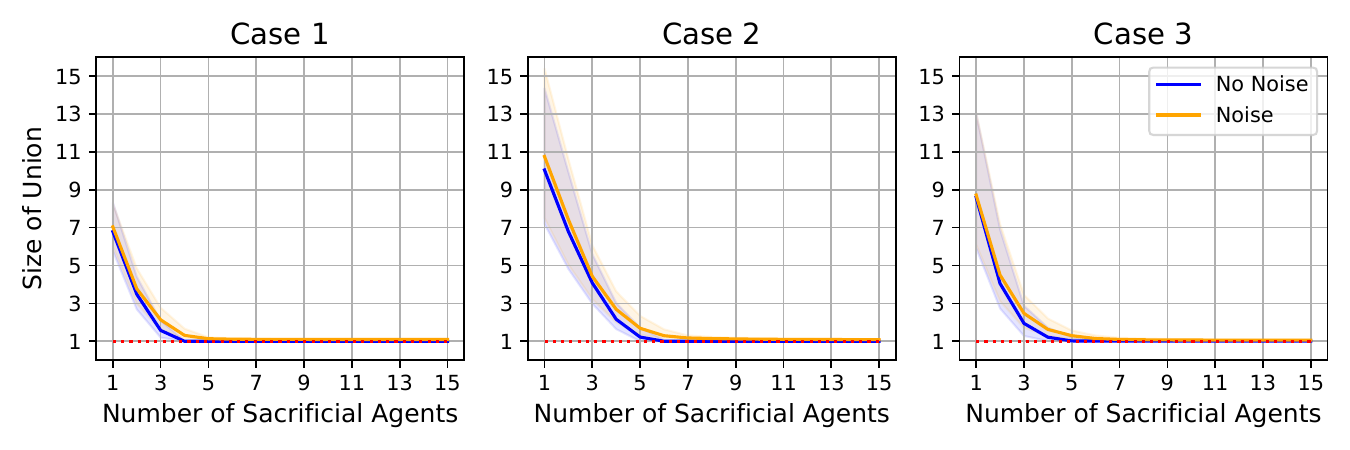}
\caption{Size of the outer union of the RRs for all feasible sets of pursuer parameters under the boundary-interception assumption.}\label{fig:union_size_boundary}
\end{figure}

Figure~\ref{fig:union_size_boundary} shows the size of the outer union of the pursuer’s RR across all feasible parameter vectors $\mathcal{T}$, normalized by the size of the true RR. The true RR area is computed by numerically integrating the RR generated from the true pursuer parameters. For each iteration, the outer union of RRs from all currently feasible parameter vectors is also numerically integrated, and the ratio of the two areas is reported. A value of one indicates perfect agreement between the union and the true RR. Blue corresponds to the noise-free case and orange to the noisy case, with shaded regions denoting the interquartile range across 500 Monte Carlo runs. The red dashed line marks the true RR size. In all cases, the union ratio decreases rapidly with the number of sacrificial agents, converging to a single consistent set after roughly four to five agents in Case 1, and five to six in Cases 2 and 3. Notably, while noisy settings reduce parameter accuracy more gradually, the overall union size still shrinks quickly, demonstrating robustness of the approach.

\begin{table}[t]
    \centering
    \begin{tabular}{lcccccc}
        \toprule
        \textbf{Coverage Threshold} & \textbf{Case 1A} & \textbf{Case 1B} & \textbf{Case 2A} & \textbf{Case 2B} & \textbf{Case 3A} & \textbf{Case 3B} \\
        \midrule
                 95\%          &  99.71\%   &  98.62\%   &  97.26\%   &  98.71\%   &  90.74\%   &  92.60\%   \\
         90\%          &  99.80\%   &  99.32\%   &  98.45\%   &  99.44\%   &  94.20\%   &  96.79\%   \\
         85\%          &  100.00\%  &  99.61\%   &  99.10\%   &  99.65\%   &  95.71\%   &  97.97\%   \\
        \bottomrule
    \end{tabular}
    \caption{Percentage of deployment steps for which the outer union of feasible RRs contained at least the indicated fraction of the true RR area.}
    \label{tab:coverage_results_bound}
\end{table}

Table~\ref{tab:coverage_results_bound} summarizes how consistently the outer union of all feasible RRs contained the true RR during learning. At each deployment step in every MC run, the ratio between the area of the true RR and the portion of that area enclosed by the current outer union was computed by numerically integrating the intersection of the true RR and outer union. A step was counted as satisfying a given coverage threshold (e.g., 95\%) if at least that fraction of the true RR area was contained within the union. The table entries show the percentage of all deployment steps across all runs for which the coverage exceeded each threshold.

The results demonstrate that in Cases~1A–2B, the learned outer union contained at least 95\% of the true RR area in more than 98\% of all evaluated steps, confirming that the set of inferred parameter vectors nearly always produced complete geometric containment of the true RR. Coverage decreases modestly in Cases~3A and~3B due to the higher-dimensional parameter inference and measurement noise, yet the union still covered at least 95\% of the true RR in roughly 91\% of all steps, showing that the algorithm maintains high geometric fidelity throughout learning.

\subsection{Interior Interception}
\begin{figure}[ht]
    \centering
\includegraphics[width=.95\linewidth,trim={0.0cm 0.cm .0cm .0cm},clip]{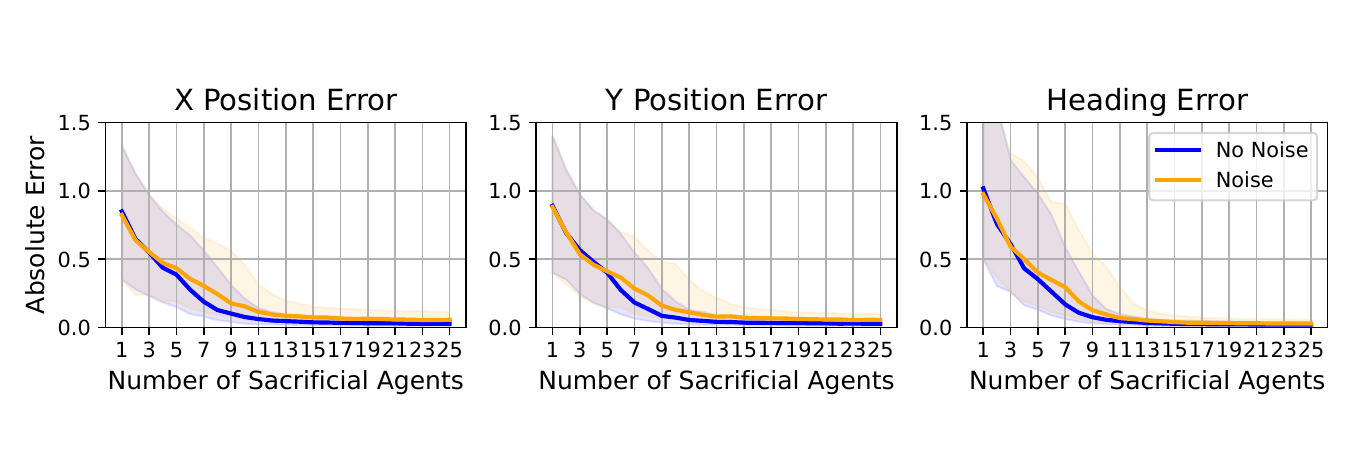}
\caption{Median absolute error (solid) and IQR (shaded) across 500 runs for Cases 1A and 1B under the interior-interception assumption. Blue indicates no measurement noise, and orange indicates noisy measurements.}\label{fig:error_known_shape_int}
\end{figure}
In this section we report results using the \assump{A1-relaxed} interior assumption. We use the same 500 MC runs to validate the learning algorithms. In general, the boundary assumptions allows the pursuer parameters to be learned faster than the interior interception assumption. This can be seen for Cases 1A and 1B with the \assump{A1-relaxed} interior assumption error plots shown in Figure~\ref{fig:error_known_shape_int}. Under these assumptions it takes around nine sacrificial agents to learn the pursuer's parameters in the noise free case (blue) and 11-12 in the noisey case (orange). It is important to note that the chart shows total number of agents deployed, not intercepted. 
Under the interior-interception case, the trajectory-selection objective intentionally favors trajectories that skim just outside the RR boundary, as these produce the strongest sensitivity of the loss to the pursuer parameters.
Consequently, many sacrificial agents are not intercepted. If surviving agents can be reused, the total number of agents required can be substantially reduced.

\begin{figure}[ht]
    \centering
\includegraphics[width=.95\linewidth,trim={0.0cm 0.cm .0cm .0cm},clip]{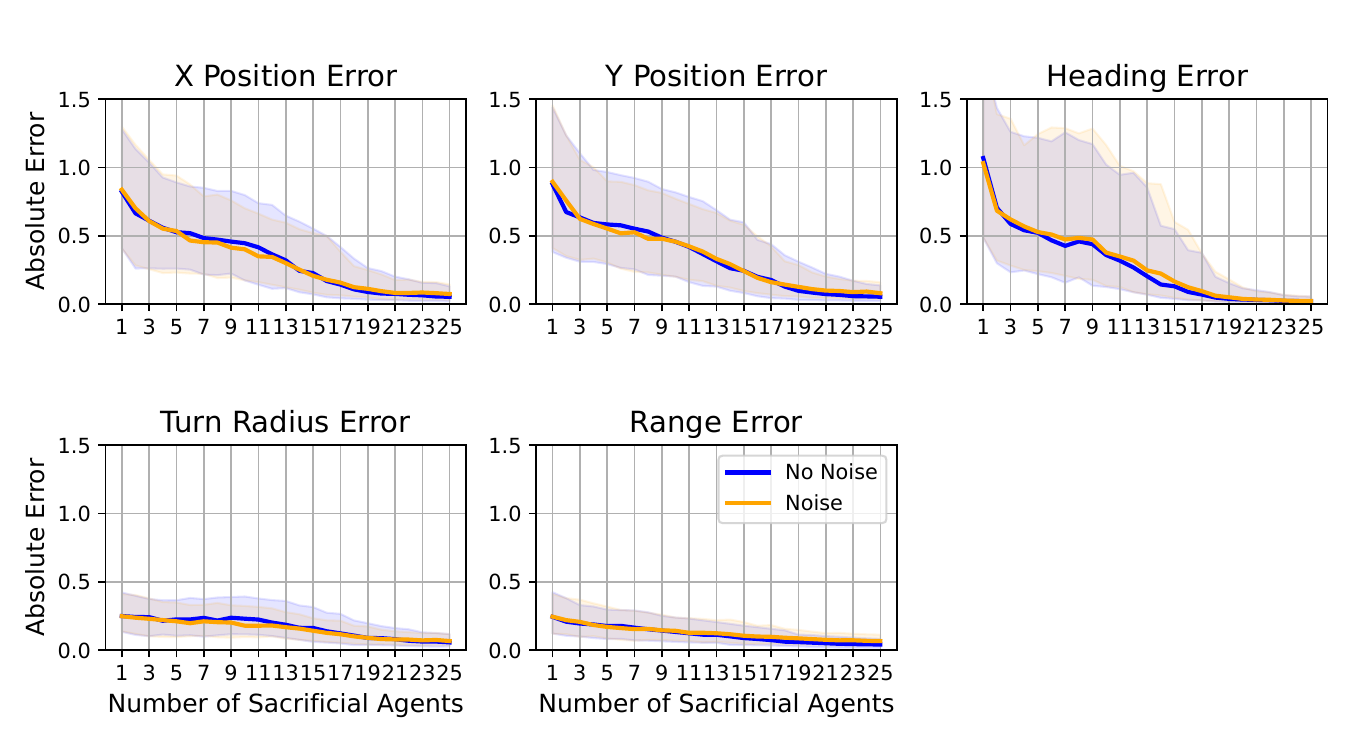}
\caption{Median absolute error (solid) and IQR (shaded) across 500 runs for Cases 2A and 2B under the interior-interception assumption. Blue indicates no measurement noise, and orange indicates noisy measurements.}\label{fig:error_known_speed_int}
\end{figure}
Figure~\ref{fig:error_known_speed_int} shows the estimation error for Cases~2A (blue) and~2B (orange) under the interior interception assumption. As shown, approximately 16–17 sacrificial agents are required for convergence in the noise-free case (blue) and 19–20 agents in the noisy case (orange). Note that the X-axis indicates the number of deployed agents, not the number intercepted. If agents can be reused after surviving an engagement, the required number of deployments would decrease accordingly. The interior-interception assumption makes each trajectory less informative than in the boundary-interception case, which is reflected in the larger number of sacrificial agents needed for convergence.

\begin{figure}[ht]
    \centering
\includegraphics[width=.95\linewidth,trim={0.0cm 0.cm .0cm .0cm},clip]{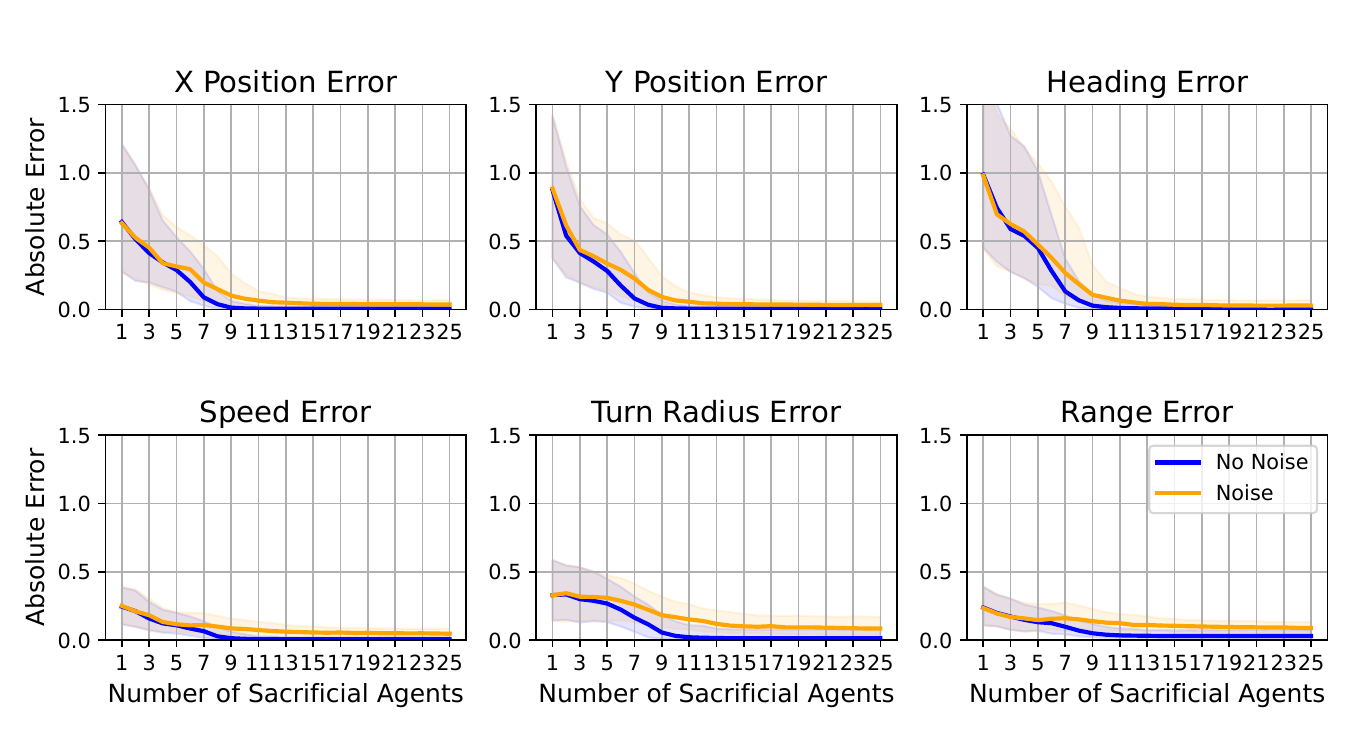}
\caption{Median absolute error (solid) and IQR (shaded) across 500 runs for Cases 3A and 3B under the interior-interception assumption. Blue indicates no measurement noise, and orange indicates noisy measurements.}\label{fig:error_unknown_speed_int}
\end{figure}

Figure~\ref{fig:error_unknown_speed_int} shows the absolute error for Cases~3A (blue) and~3B (orange) under the interior-interception assumption. Interestingly, in the noise-free case (blue) the error decreases faster than in the noise-free Case~2A (Figure~\ref{fig:error_known_speed_int}), even though more parameters are being estimated. This occurs because, under the interior-interception assumption, the binary interception outcome provides less information than observing the pursuer’s launch time. In Case~3, the launch time is assumed to be measured, and this additional information significantly accelerates error reduction. In contrast, the noisy case (orange) shows that measurement noise limits the ability to reduce uncertainty in the estimated turn radius and range. Overall, convergence occurs after approximately 7–8 sacrificial agents in the noise-free case and 10–11 agents in the noisy case.

\begin{figure}[ht]
    \centering
\includegraphics[width=.95\linewidth,trim={0.0cm 0.cm .0cm .0cm},clip]{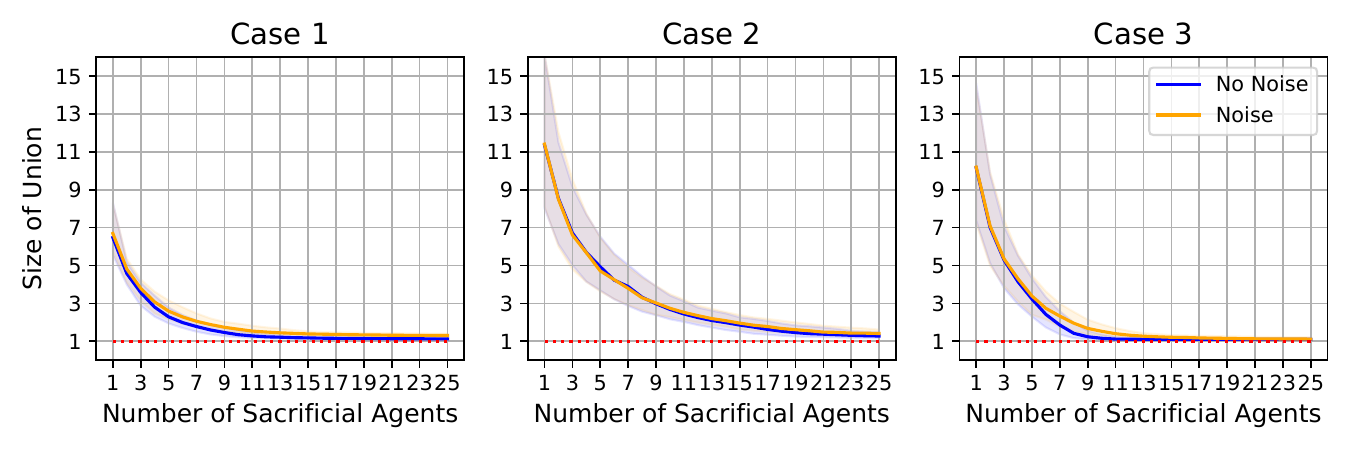}
\caption{Size of the outer union of the RR for all feasible sets of pursuer parameters under the interior-interception assumption.}\label{fig:union_size_int}
\end{figure}

Figure~\ref{fig:union_size_int} compares the area of the outer union of all feasible learned RRs to the area of the true RR, both computed via numerical integration. The results show that even though the estimation error in the noisy cases (orange) does not converge exactly to zero, the area of the outer union nevertheless approaches that of the true RR. This indicates that the learned feasible sets continue to contract toward the true reachable region, maintaining conservative geometric coverage even when the underlying parameter estimates remain imperfect.

This behavior is further quantified in Table~\ref{tab:coverage_results_int}, which reports the percentage of deployment steps across all MC runs for which the outer union of feasible RRs contained at least 95\%, 90\%, or 85\% of the true RR area. The consistently high values across all cases confirm that the learned feasible sets almost always fully covered the true reachable region throughout learning, even under noisy conditions.

\begin{table}[t]
    \centering
    \begin{tabular}{lcccccc}
        \toprule
        \textbf{Coverage Threshold} & \textbf{Case 1A} & \textbf{Case 1B} & \textbf{Case 2A} & \textbf{Case 2B} & \textbf{Case 3A} & \textbf{Case 3B} \\
        \midrule
                  95\%          &  98.72\%   &  99.05\%   &  99.42\%   &  99.09\%   &  96.66\%   &  93.07\%   \\
         90\%          &  99.67\%   &  99.70\%   &  99.76\%   &  99.71\%   &  98.24\%   &  96.11\%   \\
         85\%          &  99.85\%   &  99.91\%   &  99.89\%   &  99.82\%   &  98.98\%   &  97.98\%   \\
        \bottomrule
    \end{tabular}
    \caption{Percentage of deployment steps for which the outer union of feasible RRs contained at least the indicated fraction of the true RR area.}
    \label{tab:coverage_results_int}
\end{table}

\subsection{Path Planner}
In this section, we present results for the safe path-planning algorithm described in Section~\ref{sec:safe_path}. We performed 50 MC runs for all twelve test cases—six under the boundary-interception assumption and six under the interior-interception assumption. For each case, we report the \emph{normalized path time}, defined as the ratio between the time of a learned safe trajectory and the time of the optimal trajectory computed with perfect knowledge of the pursuer. A value of one indicates that the learning and planning algorithms together produced a trajectory as efficient as the one achievable with perfect information.

For comparison, we also include the path time obtained when no sacrificial agents are deployed. In this case, because no pursuer parameters have been learned, the only way to guarantee safety is to plan a trajectory that remains entirely outside any location that the pursuer could possibly reach. This region corresponds to the box representing all possible initial positions of the pursuer, defined by the parameter bounds $\vect{\theta}_P^{\text{min}}$ and $\vect{\theta}_P^{\text{max}}$. It is then extended by the pursuer’s range (known in Case~1) and by the maximum possible range (from $\vect{\theta}_P^{\text{max}}$) for Cases~2 and~3, to encompass any location that the pursuer could potentially reach.

\begin{figure}[ht]
    \centering
    \includegraphics[width=.95\linewidth,trim={0.0cm 0.cm .0cm .0cm},clip]{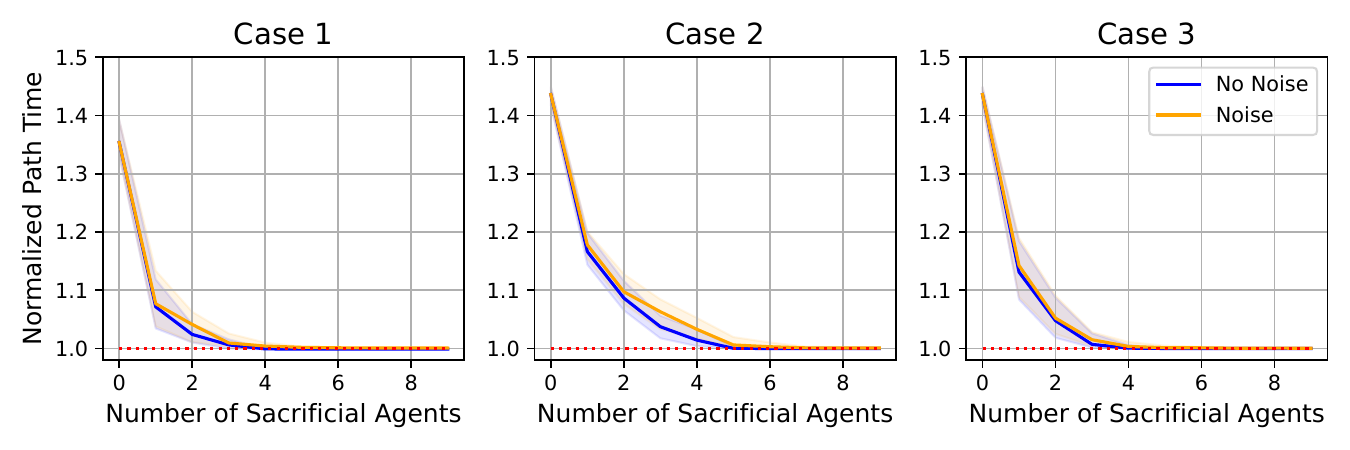}
    \caption{Median ratio between the optimal path length and the path length of the safe path planned around all feasible EZs for the boundary-interception assumption.}
    \label{fig:path_time_boundary}
\end{figure}

Figure~\ref{fig:path_time_boundary} shows the ratio between the path length of the optimal trajectory (planned with perfect knowledge) and the safe trajectory planned around all EZs generated from the current learned set of pursuer parameter vectors. The plots show the median and IQR values across 50 MC runs with varying pursuer parameters. Each boundary-interception case is shown, with the noise-free (blue) and noisy (orange) results indicated. The largest decrease in path time across all cases occurs after the first sacrificial agent, as the pursuer’s location becomes constrained to be near the observed interception point. 

In Case~1, the path length decreases more sharply after the first agent because the pursuer’s range is known, producing a circular region of radius~$R$ representing all possible pursuer locations. In Cases~2 and~3, where the range must be learned, the first sacrificial agent still yields a significant reduction in path length. As shown, the optimal path is typically achieved within three to four agents in Case~1, four to five agents in Case~2, and three to four agents in Case~3. These results demonstrate that even before the pursuer parameters are precisely learned, the safe path-planning algorithm substantially reduces path time while maintaining safety.

\begin{figure}[ht]
    \centering
\includegraphics[width=.95\linewidth,trim={0.0cm 0.cm .0cm .0cm},clip]{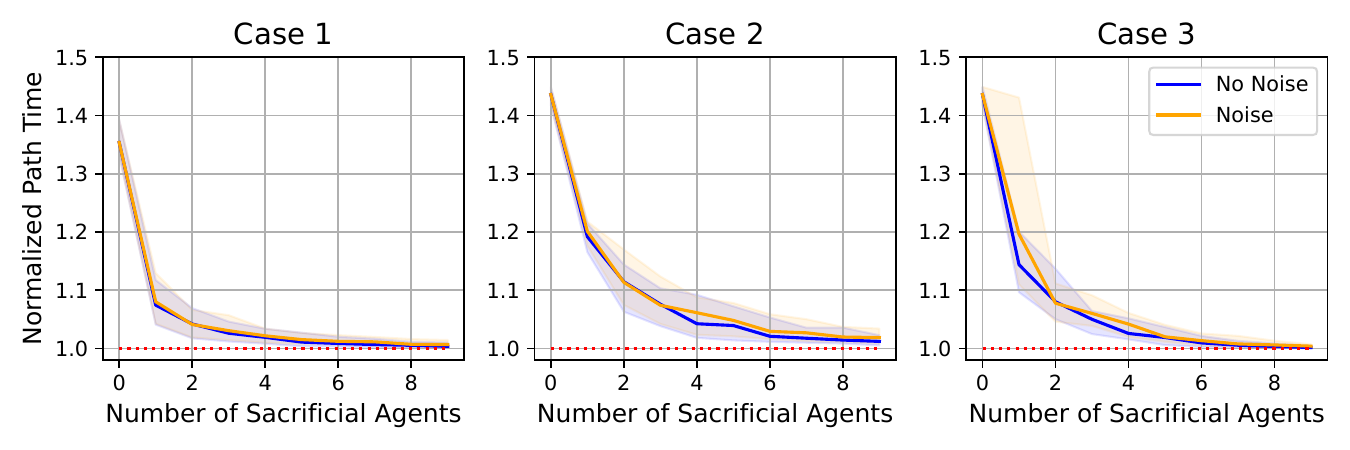}
\caption{Median ratio between the optimal path length and the path length of the safe path planned around all feasible EZs for the interior-interception assumption.}\label{fig:path_time_int}
\end{figure}

Figure~\ref{fig:path_time_int} shows the ratio between the optimal path time—computed using the true pursuer parameters—and the path time of trajectories planned using the current set of learned pursuer parameter vectors under the interior-interception assumption. Although more sacrificial agents are typically required in the interior-interception case to fully learn the pursuer parameters, the path times of the planned trajectories decrease substantially even with only a few agents.

In Case 1, the path time drops sharply with a single agent, and with two or three agents it approaches the optimal value achievable with perfect information. In Case 2, more agents are needed to achieve comparable performance, with around five or six agents required to reduce the path time to within approximately 5\% of optimal. In Case 3, where additional parameters are unknown but an extra measurement is available, similar improvements are seen with about four or five agents.

Overall, these results indicate that while many sacrificial agents may be required to fully collapse the the set of learned parameter vectors, significant performance gains can be achieved much earlier. Even a single agent can markedly reduce the safe path time, demonstrating the practical benefit of limited early learning.

\section{Conclusion}\label{sec:conclusion}
This paper presented a unified learning framework for pursuit–evasion scenarios in which autonomous agents must infer adversary capabilities from sparse interaction outcomes in order to plan safe trajectories.
% This paper presented a unified learning framework for estimating pursuer parameters from binary interception outcomes using RR models. 
The method establishes a consistent geometric foundation for interpreting sacrificial-agent outcomes by enforcing that survived trajectories remain outside the RR and intercepted ones terminate on or within it. Two formulations were developed: a boundary-interception case that assumes interception at the RR boundary and an interior-interception case that relaxes this constraint. For each, tailored loss functions were derived and optimized with gradient-based methods using automatic differentiation through the RR model.

A multi-start inference strategy was implemented using IPOPT to address the nonconvexity of the learning problem and prevent convergence to local minima. The ensemble of feasible, learned RRs progressively contracts as more sacrificial trajectories are observed.

Two complementary methods for selecting sacrificial trajectories were presented. The first is a heuristic approach that maximizes the expected spread of interception locations along the RR boundary, promoting uniform exploration of the feasible set. The second is an information-theoretic method based on Bayesian experimental design that selects the next trajectory by maximizing the $D$-score (log-determinant) of the expected Gauss–Newton information surrogate, choosing paths that most increase the curvature of the loss landscape. The heuristic method is used in the boundary-interception case, where interception locations are assumed to lie on the RR boundary and uniform boundary coverage is desirable, whereas the BED method is used in the interior-interception case, where trajectories must be selected to maximize sensitivity of the loss to model parameters.

Monte Carlo simulations over hundreds of randomized pursuer parameter vectors demonstrated that the method rapidly converges to accurate estimates even under measurement noise. In boundary-interception cases, roughly five to seven sacrificial agents were sufficient for full parameter convergence, while the interior-interception cases required slightly more trajectories but provided robustness when fewer intercepts occurred. Across all cases, the learned feasible RRs maintained over 95\% geometric coverage of the true RR in more than 94\% of all deployment steps, confirming both convergence and physical consistency of the learned models.

We also demonstrate that safe paths can be effectively planned using the learned set of pursuer parameter vectors, even before the parameter set collapses to full agreement. As shown, the use of even a single sacrificial agent can substantially reduce the safe planned path length. With additional agents, the planned paths become progressively shorter, approaching the optimal path length achievable with perfect knowledge of the pursuer parameters. This reveals a tradeoff between the number of sacrificial agents deployed the planned path length.

The proposed framework unifies geometric reachability modeling with data-driven inference, enabling pursuer characteristics to be learned directly from binary engagement outcomes. By coupling physically grounded RR models with actively selected sacrificial trajectories, the method transforms sparse interception data into actionable information about adversary capability. 

Several limitations of the present framework should be acknowledged. We assume a stationary pursuer launch point, deterministic kinematic models, and known evader behavior. The interior-interception model further assumes a simplified capture distribution along the interior of the reachable region. In addition, sensing effects such as detection uncertainty, occlusion, and adversarial adaptation are not modeled. These assumptions are intentional in order to establish a clear geometric and algorithmic foundation; relaxing them to incorporate mobility, probabilistic interception, and richer sensing models represents an important direction for future work.

Future work will extend this framework along several dimensions. First, more realistic interaction models can be incorporated by replacing the deterministic interception assumptions with probabilistic capture models that account for sensing uncertainty, seeker dynamics, and engagement stochasticity. Second, the current assumption of a stationary pursuer can be relaxed to allow mobile or adaptive adversaries, enabling learning and planning in dynamic environments. Finally, richer sacrificial-agent behaviors—such as curved trajectories, coordinated multi-agent probing, and closed-loop adaptive exploration—may substantially improve information efficiency relative to the straight-line probes considered here.

\section*{Acknowledgments}
This work was supported by the NSF IUCRC Phase I: Center for Autonomous Air Mobility and Sensing (CAAMS) under Award No.~2139551.

Portions of this manuscript were refined and edited using OpenAI’s ChatGPT model (GPT-5). The tool was employed for language polishing, LaTeX and Python code assistance, and to help organize and clarify research ideas. All technical content, algorithms, and conclusions were developed, verified, and approved by the authors.

\bibliography{sample}

\end{document}